\documentclass{ecai} 

\usepackage{latexsym}
\usepackage{amssymb}
\usepackage{amsmath}
\usepackage{amsthm}
\usepackage{booktabs}
\usepackage{enumitem}
\usepackage{graphicx}
\usepackage{color}

\usepackage{caption}     
\usepackage{subcaption}
\usepackage{bm}
\usepackage{bbm}
\usepackage{amsfonts}
\usepackage{multirow}
\usepackage{ragged2e}
\usepackage{bbold}
\usepackage{stfloats}

\newcommand{\BibTeX}{B\kern-.05em{\sc i\kern-.025em b}\kern-.08em\TeX}

\begin{document}


\begin{frontmatter}



\title{Semi-Supervised Dual-Threshold Contrastive Learning for Ultrasound Image Classification and Segmentation}


\author[A]{\fnms{Peng}~\snm{Zhang}}
\author[A]{\fnms{Zhihui}~\snm{Lai}\thanks{Corresponding Author. Email: laizhihui@szu.edu.cn}}
\author[B]{\fnms{Heng}~\snm{Kong}} 

\address[A]{College of Computer Science and Software Engineering, Shenzhen University, Shenzhen, China}
\address[B]{Department of Thyroid and Breast Surgery, Affiliated Hospital Group of Guangdong Medical University Shenzhen Baoan Central Hospital, Shenzhen, China}


\begin{abstract}
Confidence-based pseudo-label selection usually generates overly confident yet incorrect predictions, due to the early misleadingness of model and overfitting inaccurate pseudo-labels in the learning process, which heavily degrades the performance of semi-supervised contrastive learning. Moreover, segmentation and classification tasks are treated independently and the affinity fails to be fully explored. To address these issues, we propose a novel semi-supervised dual-threshold contrastive learning strategy for ultrasound image classification and segmentation, named Hermes. This strategy combines the strengths of contrastive learning with semi-supervised learning, where the pseudo-labels assist contrastive learning by providing additional guidance. Specifically, an inter-task attention and saliency module is also developed to facilitate information sharing between the segmentation and classification tasks. Furthermore, an inter-task consistency learning strategy is designed to align tumor features across both tasks, avoiding negative transfer for reducing features discrepancy. To solve the lack of publicly available ultrasound datasets, we have collected the SZ-TUS dataset, a thyroid ultrasound image dataset. Extensive experiments on two public ultrasound datasets and one private dataset demonstrate that Hermes consistently outperforms several state-of-the-art methods across various semi-supervised settings. The code is available at https://github.com/Aventador8/Hermes.
\end{abstract}
\end{frontmatter}
\section{Introduction}
In recent years, computer vision has made unprecedented advancements in natural images \cite{wu:code, wu:coarse}, and several methods \cite{bg:1, bg:3} have been successfully introduced to ultrasound images. However, these methods typically rely on large amounts of labeled data for effective model training. In the medical field, data annotation requires specialist clinical expertise, which is extremely expensive and time-consuming. Semi-supervised learning offers a promising solution by leveraging vast amounts of unlabeled data to mitigate this challenge. Recent works have explored various paradigms, such as consistency regularization \cite{bg:5}, co-training \cite{bg:7}, and self-training \cite{bg:10}. These approaches utilize unlabeled data by either constructing reliable pseudo-labels or enforcing prediction consistency under input perturbations during training. However, such approaches treat pixels or images independently, overlooking intrinsic relationships between pixels or images.\\
\begin{figure}[h]
    \vspace{-2.0em}
    \centering
    \includegraphics[width=\columnwidth]{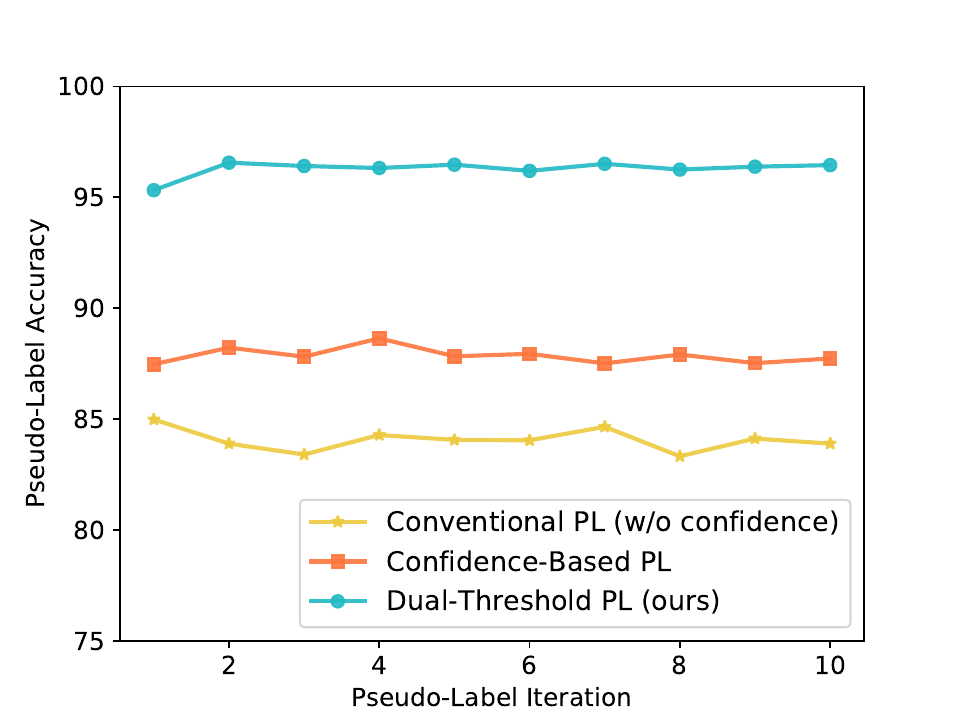}
    \vspace{0.1em}
    \caption{Comparison of pseudo-label (PL) selection accuracy between conventional PL without confidence constraint, confidence-based PL, and dual-threshold PL on CIFAR-10. Dual-threshold PL shows a substantial improvement in pseudo-label quality.}
    \label{fig:1}
    \vspace{2.0em}
\end{figure}
To capture the relationships between pixels or images, some researchers\cite{bg:13, bg:14} use contrastive learning to force the embedding representations of similar samples to be closed in the latent feature space, while the embedding representations of dissimilar samples are pushed further away. In semi-supervised settings, confidence-based pseudo-labels are used to determine the positive and negative samples for contrastive learning on unlabeled data. However, the effectiveness of semi-supervised contrastive learning heavily depends on the quality of pseudo-labels. Confidence-based pseudo-label selection frequently yields overly confident yet incorrect predictions, resulting in noisy pseudo-labels, which subsequently hinders the learning process. This phenomenon is caused by misleading during the early stages of model training and overfitting inaccurate pseudo-labels. To solve this issue, uncertainty estimation is further used to measure the uncertainty prediction of the model for a given input to avoid completely relying on the confidence value to determine the reliability of the pseudo-label. As shown in Fig. \ref{fig:1}, we surprisingly notice that, combining confidence with uncertainty, the pseudo-label accuracy is significantly improved. 

Tumor segmentation and classification are two fundamental computer vision tasks. The two tasks are closely related and share some common image features. For instance, the edge features of a tumor is important for both classification and segmentation tasks. Previous researches tend to address the two tasks independently, failing to fully exploit the potential of integrating complementary information from both tasks. Therefore, inspired by \cite{bg:15}, the two tasks can be combined to construct a joint model for improving overall performance.

In this paper, we mainly solve three questions: 1) \textit{how to improve the quality of pseudo-labels}, 2) \textit{how to make good use of high-quality pseudo-labels in two tasks} and 3) \textit{how to share information between two tasks for effective learning}. To address the first issue, we propose a novel dual-threshold pseudo-labels selection strategy, which leverages both the confidence and uncertainty to improve quality of pseudo-label. To solve the second issue, we propose a dual-threshold contrastive learning to fully use unlabeled data and capture internal relations between pixels or images. To tackle the third issue, firstly, an inter-task attention and saliency module is devised. The segmentation branch produces an attention map for the classification branch, which in turn generates a saliency map shared with the segmentation branch. This bidirectional information sharing improves performance on both tasks. Secondly, we propose an inter-task consistency learning strategy to align tumor features in high-dimensional latent space across two tasks and reduce negative transfer, improve the consistency and accuracy of the model. Additionaly, we will release a new thyroid ultrasound image dataset, termed SZ-TUS, aimed at advancing academic research.
In summary, our contributions can be described as follows:

\begin{itemize}
    \item A novel dual-threshold contrastive learning method is proposed, which can effectively alleviate the impact of noise pseudo-labels of unlabeled data.
    \item We develop the inter-task attention and saliency module particularly designed for the joint learning model, which facilitates the information sharing between ultrasound image segmentation and classification.
    \item An inter-task consistency learning strategy is devised for aligning features across segmentation and classification to reduce negative transfer.
    \item A thyroid ultrasound image dataset, which contains 982 thyroid images, comprising 766 benign and 216 malignant cases, will be public available for research purposes.
\end{itemize}


\section{Related work}
\subsection{Semi-supervised learning} The objective of semi-supervised learning is to leverage the unlabeled data to provide additional information during training. Semi-supervised learning mainly consists of pseudo-label prediction, entropy minimization and consistency regularization. Pseudo-label prediction methods \cite{rw:1} train a model on labeled data and then use the trained model to generate pseudo-labels for unlabeled data. Entropy minimization \cite{rw:6} encourages confident predictions by steering the model to minimize the entropy of the prediction probability distribution, ensuring that high-quality predictions exhibit low entropy. Consistency regularization \cite{rw:4, rw:5} enforces that predictions remain consistent across different views or perturbations of the same image. Building upon consistency regularization, we propose a novel inter-task consistency learning strategy to align tumor features across two tasks, avoiding negative transfer.

\subsection{Contrastive learning} Contrastive learning aims to minimize the distance between positive pairs and maximize the distance between negative pairs in latent space. SimCLR \cite{rw:7} and  MoCo \cite{rw:8} propose pipelines for learning augmentation-invariant in self-supervised visual representations, which demonstrate superior performance. Recent progress in this field focus on the use of contrastive learning in semi-supervised scenarios \cite{rw:9, rw:11}, allowing model pre-trained on classification task to generalize effectively to segmentation task. However, as illustrated in Fig. \ref{fig:1}, relying solely on confidence for pseudo-label selection often leads to low accuracy, which results in noisy sampling during the construction of sample pairs and degrades the performance of contrastive learning. To address this issue, we propose a novel dual-threshold approach that combines uncertainty and confidence to improve the accuracy of pseudo-labels, thereby enhancing the robustness and effectiveness of contrastive learning.

\subsection{Uncertainty estimation} Uncertainty estimation in network predictions has been the subject of significant research, with several common techniques emerging \cite{rw:12, rw:13}. These include: 1) leveraging bayesian neural network to model the probability distribution of model parameters \cite{rw:15}, 2) assessing the variance across multiple predictions of the same input under varying perturbations \cite{rw:16}, and 3) evaluating the variance among different predictions for the same input \cite{rw:17}. However, these methods have high computational complexity. In our method, uncertainty is estimated by calculating the entropy of probability distribution obtained from weakly augmented input images to reduce computation.


\section{Method}
\begin{figure*}
    \centering
    \includegraphics[width=\linewidth]{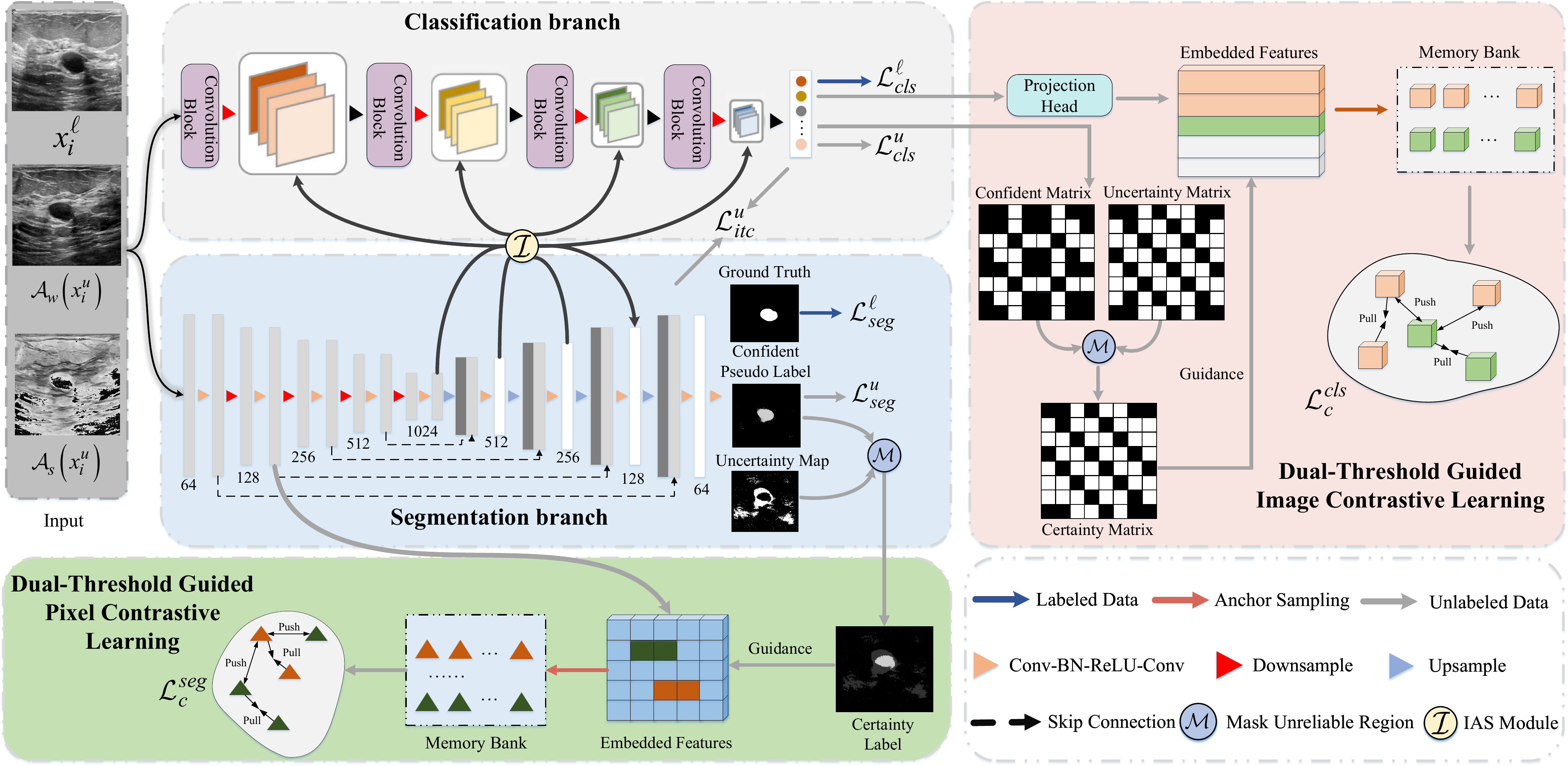}
    \vspace{0.001em}
    \caption{The proposed Hermes consists of two branches, each branch leveraging dual-threshold contrastive learning to enhance the performance. The $\mathcal{I}$ symbol enclosed in a circle corresponds to the Inter-task Attention and Saliency (IAS) Module, which can be found in Fig.~\ref{fig:3}. Colored arrows indicate the distinct processing flows for different data streams, illustrating how various components interact within network.}
    \label{fig:2}
    \vspace{1.5em}
\end{figure*}
Confidence-based pseudo-label selection often generates inaccurate pseudo-labels, due to the misleadingness of early model and overfitting incorrect pseudo-labels. Moreover, previous researches tend to solve segmentation and classification task independently, failing to fully explore complementary informations between two tasks to improve performance of the model. Therefore, we propose a novel dual-threshold pseudo-label learning strategy to enhance quality of pseudo-labels. To address second issue, a joint Hermes\footnote{Hermes is the messenger god and protector in Greek mythology, and this dual identity is similar to the duality in our work.} model is designed, as shown in Fig. \ref{fig:2}. Additionally, to avoid negative transfer, an inter-task consistency learning strategy is proposed to align tumor features across two tasks.


\subsection{Supervised segmentation and classification}
\textbf{Pipeline rationality.} A common backbone assumes the equivalence and effectiveness of learned features across tasks, but improper feature sharing can cause negative transfer, degrading performance on some tasks \cite{bd:19}. Therefore, we use two different backbones to avoid gradients conflict. 
\\
\textbf{Pixel-level segmentation.} Given a labeled dataset $D_L\! = \left \{\left(x_i^{\ell},y_i^{\ell},z_i^{\ell} \right ) ,i=1,\dots ,N \right \} $, where $x_i^{\ell}$ denotes original image, $y_i^{\ell}$ denotes the pixel-wise annotation map and $z_i^{\ell}$ denotes image-level classification label. For a given image $x_i^{\ell}$, we use $y_i^{\ell}$ to calculate the supervised segmentation loss:
\begin{equation}
\mathcal{L}_{seg}^{\ell}=\frac{1}{2\left | D_{L} \right | }\! \sum_{i=1}^{\left | D_{L} \right|}\!\left ( \mathcal{L}_{CE}\!\left ( f_{\theta}^{s}\!\left ( x_{i}^{\ell} \right )\!,\!y_{i}^{\ell}  \right )\! +\! \mathcal{L}_{Dice}\!\left ( f_{\theta}^{s}\!\left ( x_{i}^{\ell} \right )\!,\!y_{i}^{\ell}  \right ) \right ) 
\end{equation}
where $\mathcal{L}_{CE}$ and $\mathcal{L}_{Dice}$ denote the cross-entropy loss and Dice loss, respectively, $\left | \cdot \right |$ represents set length. $f_{\theta}^{s}\left ( \cdot \right )$ refers to the predictions generated by the segmentation branch.
\\
\textbf{Image-level classification.} The calculation of classification branch loss is analogous to segmentation branch. We use cross-entropy to calculate supervised classification loss:
\begin{equation}
\mathcal{L}_{cls}^{\ell} = \frac{1}{\left | D_{L} \right | } \sum_{i=1}^{\left | D_{L} \right |} \mathcal{L}_{CE}\left ( f_{\theta }^{c}\left ( x_{i}^{\ell} \right ) \!,\!z_{i}^{\ell} \right )     
\end{equation}
where $f_{\theta}^{c}\left ( \cdot \right )$ represents the output prediction from classification branch.

\subsection{Inter-task attention and saliency module}
\begin{figure*}
    \centering
    \includegraphics[width=\linewidth]{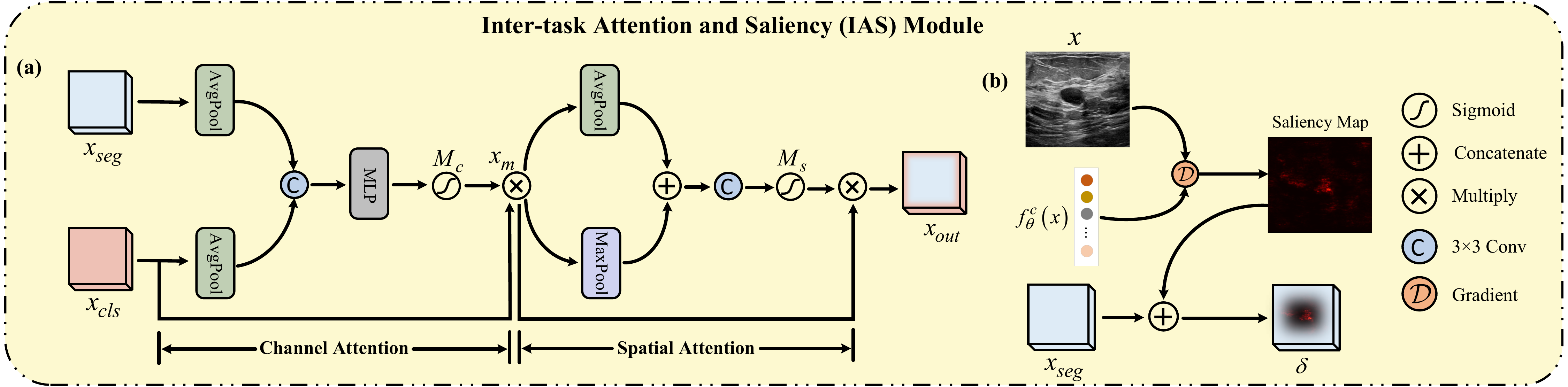}
    \vspace{0.001em}
    \caption{Overall workflow of inter-task attention and saliency module. (a) The channel attention utilizes average-pooling to aggregate spatial information. The spatial attention utilizes max-pooling and average-pooling to extract main target object and the range of the target object, respectively. (b) The classification function generates a first-order expansion of the input image, producing a saliency map that highlights key features and gradients. }
    \label{fig:3}
    \vspace{1.0em}
\end{figure*}

To facilitate information sharing between two tasks, we propose a novel inter-task attention and saliency module particularly designed for the joint learning model.
\\
\textbf{Inter-task attention.}  The inputs of classification branch include two parts: $x_{cls}$ from the classification branch features and $x_{seg}$ derived from the segmentation branch features. A novel inter-task attention mechanism is designed to direct classification branch focusing effectively on tumor, as shown in Fig. \ref{fig:3}$\left(a\right)$, thereby improving the overall classification accuracy.

Channel attention mechanism is designed to identify the most relevant information for categorization task. By employing average pooling to aggregate spatial information, the model can effectively capture the extent of the target object \cite{bd:1}. Therefore, we use average pooling to devise channel attention mechanism to capture the size of tumor. In short, the channel attention mechanism is computed as:
\begin{equation}
    \begin{aligned}
    M_c&=\sigma \left ( MLP\left ( AvgPool\left ( f^{3\times3} \left [ x_{cls};x_{seg} \right ] +b_c \right )  \right )  \right ) \\
    &=\sigma \left ( W_1\left ( W_0\left ( AvgPool\left ( W_c\cdot \left [ x_{cls};x_{seg} \right ] +b_c \right )  \right )  \right )  \right ) 
    \end{aligned}
    \label{eq:3}
\end{equation}
where $\left [ x_{cls};x_{seg} \right ]$ denotes the channel-wise concatenation of the classification and segmentation features, $f^{3\times3}$ represents a convolution operation with a filter size of $3\times3$, and $\sigma$ is the sigmoid activation function. The weight matrix $W_c$ corresponds to the $3\!\times\!3$ convolution layer, with $b_c$ being its associated bias. Additionally, $W_0$ and $W_1$ represent the weights of the multi-layer perceptron (MLP), where a ReLU activation follows $W_0$.

Spatial attention mechanism determines where key informative regions are located, which complements the channel attention by providing a comprehensive understanding of both the content and spatial distribution of key features. Max-pooling collects additionally crucial information about distinctive object area, allowing for more precise channel-wise attention inference \cite{bd:2}. Thus, combining average pooling with max-pooling enhances the overall accuracy of feature recognition. In summary, the spatial attention mechanism is computed as follows:
\begin{equation}
    \begin{aligned}
        M_s&=\sigma \left ( f^{3\times 3}\left ( \left [ AvgPool\left ( x_{m} \right ) ;MaxPool\left ( x_{m} \right )  \right ]  \right )  \right ) \\
        &=\sigma \left ( W_s\cdot \left [ AvgPool\left ( x_{m} \right ) ;MaxPool\left ( x_{m} \right )  \right ] +b_s \right ) 
    \end{aligned}
    \label{eq:4}
\end{equation}
where $W_{s}$ represents the weight of $3\times3$ convolution with $b_{s}$ being its bias, $x_m$ represents the output of the channel attention mechanism.

The final output of inter-task attention modules can be written as:
\begin{equation}
    \begin{aligned}
        x_m&=M_c\otimes x_{cls} \\
        x_{out}&=M_s\otimes x_{m}
    \end{aligned}
\end{equation}
where $\otimes$ denotes element-wise multiplication, and $x_{out}$ represents the final output of inter-task attention mechanism. 
\\
\textbf{Saliency module.}  The saliency map is utilized to highlight the most important features and regions within an image that are crucial for classification \cite{bd:4}, thereby enhancing interpretability of the prediction of model. Furthermore, saliency map is similar to segmentation mask, as shown in Fig. \ref{fig:3}$\left(b\right)$, making it logical to guide segmentation:
\begin{equation}
\delta = \mathcal{D}\left ( x, f_{\theta}^{c}\left ( x \right ) \right )  \oplus x_{seg}
\end{equation}
where $\mathcal{D}\left( \cdot\right)$ represents the derivative of classification prediction with respect to input image $x$, $\oplus$ denotes the channel-wise concatenation.
\subsection{Dual-threshold pseudo-label selection}
To prevent models from training with inaccurate pseudo-labels, it is crucial to improve the accuracy of pseudo-labels. Thus, we propose a novel dual-threshold pseudo-label learning to improve the quality of pseudo-labels.
\\
\textbf{Dual-threshold pseudo-label learning.} Confidence-based selection reduces pseudo-label error rates, compared to without using confidence, but the misleadingness of early model and overfitting inaccurate pseudo-labels limit its effectiveness, as shown in Fig. \ref{fig:1}. Thus, a novel dual-threshold pseudo-label learning is proposed, combining confidence and uncertainty, which effectively improve the quality of pseudo-labels. Given an unlabeled dataset $D_U\!=\left \{ x_j^u, j=1,\dots ,M\right \}$, where $M\!\gg\!N$. The formula is as follows:
\begin{equation}
    w=\mathbb{1}\!\left [ u\!\left ( f_{\theta}\!\left ( \mathcal{A}_{w}\!\left ( x_{j}^{u} \right )  \right ) \right )\! \le \kappa \right ]\cdot  \mathbb{1}\!\left [ max\!\left ( f_{\theta}\!\left ( \mathcal{A}_{w}\!\left ( x_{j}^{u} \right ) \right)\right)\! \ge \tau  \right ]
\end{equation}
\begin{equation}
    u\left ( p\right )= - \sum_{j=1}^{M} p log\left ( p +\epsilon  \right )   
\end{equation}
\\
where $\mathcal{A}_{w}$ denotes the weakly augmented image, $u\left ( p\right )$ represents the uncertainty of predictions. $f_{\theta}\left(\cdot\right)$ represents the output predictions of classification or segmentation branch. $\kappa$ and $\tau$ refer to uncertainty and confidence thresholds, respectively, while $\epsilon$ is a small constant introduced to prevent singularities.
\\
\textbf{Unsupervised loss calculation.} To maintain consistency, the predictions of the strongly augmented images should be consistent with the weakly augmented images. However, since inaccurate pseudo-labels are not suitable for supervision, dual-threshold pseudo-label learning is used to filter out inaccurate pseudo-labels. In this process, the weakly augmented images $\mathcal{A}_{w}\left(x_{j}^{u}\right)$ are utilized to supervise the predictions of strongly augmented images:
\begin{equation}
\mathcal{L}_{seg}^{u}\!=\frac{1}{\left | D_{U}\right|}\!\sum_{j=1}^{\left|D_{U} \right|}\!w\cdot\!\mathcal{L}_{Dice}\!\left ( f_{\theta }^{s}\left ( \mathcal{A}_{s}\!\left ( x_{j}^{u} \right ) \right )\!,\!f_{\theta}^{s}\left ( \mathcal{A}_{w}\!\left ( x_{j}^{u} \right )  \right )  \right )   
\end{equation}
\begin{equation}
        \mathcal{L}_{cls}^{u} =\frac{1}{\left | D_{U} \right | }\!\sum_{j=1}^{\left | D_{U} \right |}\!w\cdot \mathcal{L}_{CE}\!\left ( f_{\theta }^{c}\left ( \mathcal{A}_{s}\left ( x_{j}^{u} \right ) \right )\!,\!f_{\theta}^{c}\left ( \mathcal{A}_{w}\left ( x_{j}^{u} \right )  \right )  \right )     
\end{equation}
where $\mathcal{A}_{s}\left(x_{j}^{u}\right)$ denotes the strongly augmented image.
\subsection{Dual-threshold contrastive learning}
Cross-entropy loss treats pixels or images independently, neglecting their intrinsic relationships. Thus, we utilize contrastive learning mechanism to capture this relation. To alleviate the impact of noisy pseudo-labels for contrastive learning, we use aforementioned dual-threshold learning to select anchors, where pixels or images with higher confidence and lower uncertainty are selected.
\\
\textbf{Pixel contrastive learning.}  To calculate the contrastive loss, we adopt the widely-used InfoNCE \cite{bd:3} loss function. Let the embedding vector of a pixel be represented by $i$, and its corresponding semantic label by $\tilde{c}$. Positive samples are defined as those pixels that also belong to class $\tilde{c}$, while negative samples consist of pixels belonging to other categories. The dual-threshold pixel contrastive loss use aforementioned dual-threshold learning to filter out unreliable pixels, as follow:
\begin{equation}
    \mathcal{L}_{c}^{seg}\!=-\frac{1}{\left|\mathcal{P}_i\right|}\!\sum_{i^{+}\in\mathcal{P}_i}^{}\!\!log\frac{w\cdot\text{exp} \!\left ( i\!\cdot\! i^{+}\!/\tau\right) }{\text{exp}\! \left ( i\!\cdot\! i^{+}\!/\tau  \right )+ {\textstyle \sum_{i^{-}\in \mathcal{N}_{i} }^{}} \text{exp} \!\left ( i\!\cdot\! i^{-}\!/\tau \right ) }  
\end{equation}
where $\mathcal{P}_i$ and $\mathcal{N}_i$ denote the set of pixel embeddings for positive and negative samples, respectively. $i^{+}$ refers to positive sample, while $i^{-}$ refers to negative sample. The temperature constant $\tau$ is set to 0.07. Notably, the positive or negative samples and the anchor $i$ are not required to be from the same image.
\\
\textbf{Image contrastive learning.} Class-aware contrastive learning incorporates class-specific information for unlabeled data while mitigating the effects of noisy samples. This method improves the model's ability to accurately distinguish between different classes, even when labels are unavailable. Leveraging the pseudo-labels, image embeddings $v$ from same category are considered positive pairs, rather than those from same image. Additionally, we also use aforementioned dual-threshold learning to filter out inaccurate pseudo-labels. The dual-threshold guided image-level contrastive loss is formulated as: 
\begin{equation}
    \mathcal{L}_{c}^{cls}\!=\!-\frac{1}{\left | P \right | }\!\sum_{p\in P}^{}\!log\frac{w_{ij}\cdot\!\text{exp}\left ( v_i\cdot v_p /\tau  \right )  }{ {\textstyle \sum_{j=1}^{2N}\mathbb{1}_{j\not{=}i }\text{exp} \left ( v_i\cdot v_j/\tau  \right )  } }  
\end{equation}
\begin{equation}
    w_{ij}=
    \begin{cases}
        1 & \text{if $j=i$} \\
        w & \text{otherwise}
    \end{cases}
\end{equation}
where $P$ denotes the set of image embeddings for positive samples. $v_p$ is the positive sample of corresponding image anchor $v_i$. $i$ and $j$ are indices of $v$ for augmented sample.
\\
\textbf{Anchor sampling.} The certainty pseudo-labels and certainty matrix of unlabeled images are used as the basis for contrastive learning. The certainty matrix and certainty labels represent pseudo-labels with high confidence and low uncertainty. The embedded features corresponding to these certainty pseudo-labels and certainty matrix are used for contrastive learning. Pixels or images with inconsistent predictions between weakly and strongly augmented views are considered hard samples, while those with consistent predictions are considered simple samples. We randomly sample n/2 pixels or images from each set (hard and simple samples). To reduce the cost of pixel contrastive learning, Anchor sampling is conducted on feature maps with a 4x downsampling ratio.
\\
\textbf{Memory bank.} The efficacy of contrastive learning is closely tied to the amount of contrastive negative samples. To mitigate this, a fixed-size memory buffer is constructed, which records sampled negative samples, and this buffer is updated with each iteration. 
\subsection{Inter-task consistency learning strategy} 
In order to avoid negative transfer for reducing features discrepancy between two tasks, we further propose an inter-task consistency learning strategy. By pulling classification and segmentation features closer together in high-dimensional latent space, it makes them more suitable for sharing across tasks. To achieve this, we adopt a normalized $\mathbb{L}_2$ consistency loss, which is equivalent to $1 - \text{cosine similarity}$, between the classification and segmentation features, as follows:
\begin{equation}
    \mathcal{L} _{itc}^{u}=1-\frac{F_{cls}^{T}F_{seg}}{\left \| F_{cls} \right \|_{2} \left \| F_{seg} \right \|_{2} } 
\end{equation}
where $F_{cls}$ and $F_{seg}$ denote the final output features of the classification encoder (before softmax) and the last layer features of the segmentation encoder, respectively, and are then each projected using a separate projection head.
\subsection{The overall objective functions}
By combining the dual-threshold pseudo-label selection strategy with consistency learning, we propose a novel hybrid semi-supervised approach to boost the overall performance of the Hermes model in ultrasound image segmentation and classification tasks. The total loss function used to train the model, which integrates both labeled and unlabeled data, is defined as follows:
\begin{equation}
    \begin{aligned}
    \mathcal{L}_{total}=&\lambda_{seg}\left ( \mathcal{L}_{seg}^{\ell}+\mathcal{L}_{seg}^{u}+\alpha\mathcal{L}_{c}^{seg}   \right ) \\
    &+\lambda_{cls}\left ( \mathcal{L}_{cls}^{\ell}+\mathcal{L}_{cls}^{u}+\beta\mathcal{L}_{c}^{cls}\right)+\gamma \mathcal{L}_{itc}^{u}  
    \end{aligned}
\end{equation}
\\
where $\lambda_{seg}$ and $\lambda_{cls}$ are the trade-off factors to balance the contribution between segmentation and classification. $\alpha$, $\beta$ and $\gamma$ are the weights of segmentation contrastive loss $\mathcal{L}_{c}^{seg}$, classification contrastive loss $\mathcal{L}_{c}^{cls}$ and consistency loss $\mathcal{L}_{itc}^{u}$, respectively, as stated in previous subsections.


\section{Experiments and results}
\subsection{Experimental Setup}
\textbf{Datasets.}  We validate our Hermes on two public breast ultrasound image dataset and one private thyroid ultrasound image dataset, which will be publicly available when this paper is accepted.
\begin{itemize}
    \item \textbf{BUSI Dataset}\cite{bd:5} includes 780 breast ultrasound images, comprising 133 normal images, 437 benign images and 210 malignant images. The experiments are conducted using a subset of 647 abnormal images, where 72, 144 and 216 images are labeled. 
    \item \textbf{UDIAT Dataset}\cite{bd:6} includes 163 breast ultrasound images, comprising 109 benign and 54 malignant cases. According to different semi-supervised settings, 20, 40 and 60 images are labeled.
    \item \textbf{SZ-TUS Dataset} contains 982 thyroid ultrasound images, comprising 766 benign images and 216 malignant images. We randomly select 60 and 120 images as labeled images.
\end{itemize}
All images are resized to \(224 \times 224\) pixels. Each dataset is randomly divided into training and validation sets at a ratio of 7:3.
\\
\textbf{Evaluation metrics.}  To evaluate the performance of tumor segmentation task, the Dice Coefficient (Dice) is employed as primary evaluation metric, accompanied by the standard deviation (std) of the Dice score for robustness assessment. For the classification task, model performance is measured using accuracy as the evaluation metric.
\\
\textbf{Implementations details.}  The segmentation branch employs a pretrained U-Net\cite{bd:7} as its backbone, while the classification branch utilizes a pretrained ResNet-18\cite{bd:18}. Identical training configurations are applied across all datasets. The classification branch is optimized using SGD with a weight decay of 0.0001 and momentum of 0.9, while the segmentation branch uses the ADAM optimizer. Both branches have an initial learning rate of 0.0001, which decays following a polynomial strategy: $lr=lr_{init} \cdot \left( 1-\frac{iter}{iter_{total}}  \right)^{0.9}$. The mini-batch size is set to 8, with the number of labeled and unlabeled images equal. Weak data augmentation techniques, such as random horizontal flipping, random scaling, and random cropping, are applied to both labeled and unlabeled images. In addition to these, strong augmentations include blurring and color jittering. The task tradeoff hyperparameters are set as $\lambda_{cls}=1.0$ and $\lambda_{seg}=0.5$, while the consistency loss hyperparameter is configured with $\gamma=0.1$. For contrastive learning, the weights are assigned as $\alpha=0.8$ and $\beta=0.2$. The feature dimensions are set to 128 for image-level contrast and consistency learning, and 256 for pixel-level contrast.

\subsection{Comparison with state-of-the-art methods}
\textbf{Segmentation result.}  We compare Hermes against six recent semi-supervised segmentation approaches. CPS\cite{bd:8}, ABD\cite{bd:9} and SS-Net\cite{bd:10} are primarily based on consistency learning. DiffRect\cite{bd:11} utilizes a diffusion model. AD-MT\cite{bd:12} is based on the teacher-student framework. DCNet\cite{bd:13} employs decoupled consistency. For a fair comparison, all methods are evaluated using a U-Net backbone under the same experimental conditions.

As shown in Tab. \ref{tab:1}, we evaluate Hermes on three datasets using the Dice metric. The first row reports the performance of the baseline model trained only on labeled data. Across all partitioning protocols, Hermes demonstrates significant average Dice improvement of approximately 5\%. Specifically, on the BUSI dataset, under the 216, 144, and 72 label partition protocols, our method surpasses the baseline (SupOnly) by +4.62\%, +6.47\%, and +5.35\%, respectively. Notably, with 72 labels, our method exceeds the performance of the previous state-of-the-art approach ABD\cite{bd:9} by +2.37\%. Furthermore, on the UDIAT dataset, our method delivers superior Dice scores compared to the latest semi-supervised segmentation method AD-MT\cite{bd:12} with gains of +2.28\%, +3.12\%, and +3.34\% under the 60, 40, and 20 label settings, respectively. Finally, on the private SZ-TUS dataset, Hermes consistently outperforms ABD \cite{bd:9}, achieving improvements of +1.69\% and +1.71\% under the 120-label and 60-label settings, respectively.

\begin{figure*}[t]
    \centering
    \begin{minipage}[t]{0.08\linewidth}   
        \centering
        \includegraphics[scale=0.2]{Figure/comparison/original_image/val_163_input.png}
    \end{minipage}
    \hspace{0.1em}
    \begin{minipage}[t]{0.08\linewidth}
        \centering
        \includegraphics[scale=0.2]{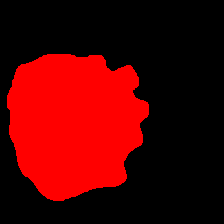}
    \end{minipage}
    \hspace{0.1em}
    \begin{minipage}[t]{0.08\linewidth}
        \centering
        \includegraphics[scale=0.2]{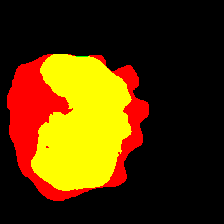}
    \end{minipage}
    \hspace{0.1em}
    \begin{minipage}[t]{0.08\linewidth}
        \centering
        \includegraphics[scale=0.2]{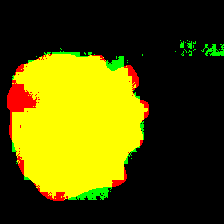}
    \end{minipage}
    \hspace{0.1em}
    \begin{minipage}[t]{0.08\linewidth}
        \centering
        \includegraphics[scale=0.2]{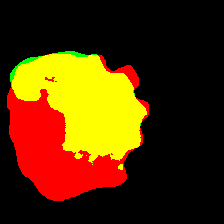}
    \end{minipage}
    \hspace{0.1em}
    \begin{minipage}[t]{0.08\linewidth}
        \centering
        \includegraphics[scale=0.2]{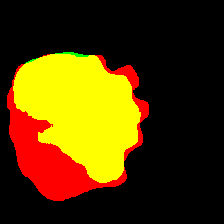}
    \end{minipage}
    \hspace{0.1em}
    \begin{minipage}[t]{0.08\linewidth}
        \centering
        \includegraphics[scale=0.2]{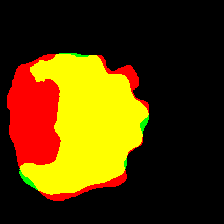}
    \end{minipage}
    \hspace{0.1em}
    \begin{minipage}[t]{0.08\linewidth}
        \centering
        \includegraphics[scale=0.2]{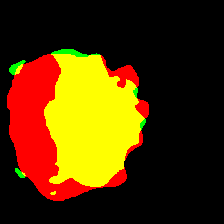}
    \end{minipage}
    \hspace{0.1em}
    \begin{minipage}[t]{0.08\linewidth}
        \centering
        \includegraphics[scale=0.2]{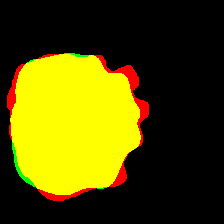}
    \end{minipage}
    \par   
    \begin{minipage}[t]{0.08\linewidth}   
        \centering
        \includegraphics[scale=0.2]{Figure/comparison/original_image/val_178_input.png}
    \end{minipage}
    \hspace{0.1em}
    \begin{minipage}[t]{0.08\linewidth}
        \centering
        \includegraphics[scale=0.2]{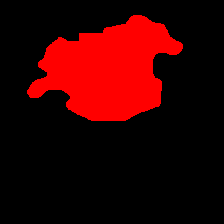}
    \end{minipage}
    \hspace{0.1em}
    \begin{minipage}[t]{0.08\linewidth}
        \centering
        \includegraphics[scale=0.2]{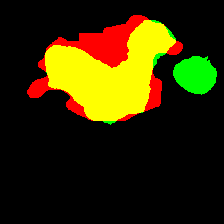}
    \end{minipage}
    \hspace{0.1em}
    \begin{minipage}[t]{0.08\linewidth}
        \centering
        \includegraphics[scale=0.2]{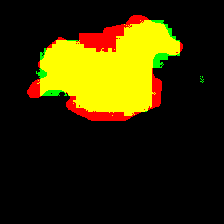}
    \end{minipage}
    \hspace{0.1em}
    \begin{minipage}[t]{0.08\linewidth}
        \centering
        \includegraphics[scale=0.2]{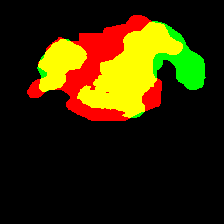}
    \end{minipage}
    \hspace{0.1em}
    \begin{minipage}[t]{0.08\linewidth}
        \centering
        \includegraphics[scale=0.2]{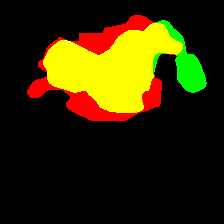}
    \end{minipage}
    \hspace{0.1em}
    \begin{minipage}[t]{0.08\linewidth}
        \centering
        \includegraphics[scale=0.2]{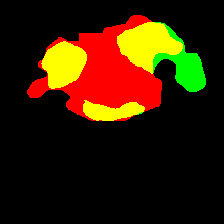}
    \end{minipage}
    \hspace{0.1em}
    \begin{minipage}[t]{0.08\linewidth}
        \centering
        \includegraphics[scale=0.2]{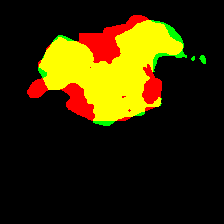}
    \end{minipage}
    \hspace{0.1em}
    \begin{minipage}[t]{0.08\linewidth}
        \centering
        \includegraphics[scale=0.2]{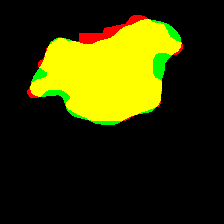}
    \end{minipage}
    \par   
    \begin{minipage}[t]{0.08\linewidth}   
        \centering
        \includegraphics[scale=0.2]{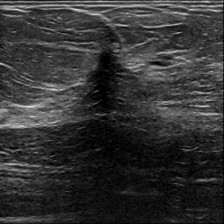}
    \end{minipage}
    \hspace{0.1em}
    \begin{minipage}[t]{0.08\linewidth}
        \centering
        \includegraphics[scale=0.2]{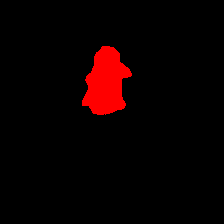}
    \end{minipage}
    \hspace{0.1em}
    \begin{minipage}[t]{0.08\linewidth}
        \centering
        \includegraphics[scale=0.2]{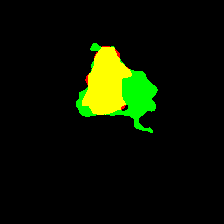}
    \end{minipage}
    \hspace{0.1em}
    \begin{minipage}[t]{0.08\linewidth}
        \centering
        \includegraphics[scale=0.2]{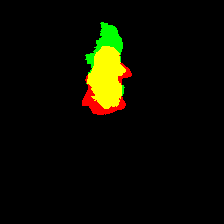}
    \end{minipage}
    \hspace{0.1em}
    \begin{minipage}[t]{0.08\linewidth}
        \centering
        \includegraphics[scale=0.2]{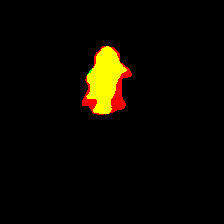}
    \end{minipage}
    \hspace{0.1em}
    \begin{minipage}[t]{0.08\linewidth}
        \centering
        \includegraphics[scale=0.2]{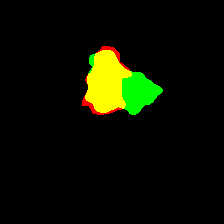}
    \end{minipage}
    \hspace{0.1em}
    \begin{minipage}[t]{0.08\linewidth}
        \centering
        \includegraphics[scale=0.2]{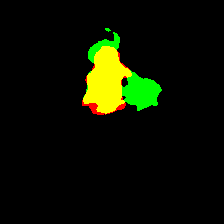}
    \end{minipage}
    \hspace{0.1em}
    \begin{minipage}[t]{0.08\linewidth}
        \centering
        \includegraphics[scale=0.2]{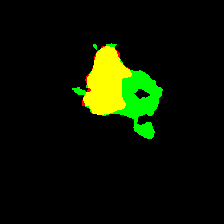}
    \end{minipage}
    \hspace{0.1em}
    \begin{minipage}[t]{0.08\linewidth}
        \centering
        \includegraphics[scale=0.2]{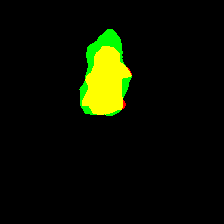}
    \end{minipage}
    \par   
    \begin{minipage}[t]{0.08\linewidth}   
        \centering
        \includegraphics[scale=0.2]{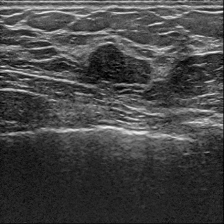}
    \end{minipage}
    \hspace{0.1em}
    \begin{minipage}[t]{0.08\linewidth}
        \centering
        \includegraphics[scale=0.2]{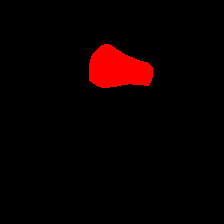}
    \end{minipage}
    \hspace{0.1em}
    \begin{minipage}[t]{0.08\linewidth}
        \centering
        \includegraphics[scale=0.2]{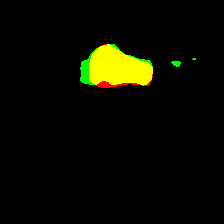}
    \end{minipage}
    \hspace{0.1em}
    \begin{minipage}[t]{0.08\linewidth}
        \centering
        \includegraphics[scale=0.2]{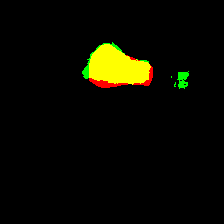}
    \end{minipage}
    \hspace{0.1em}
    \begin{minipage}[t]{0.08\linewidth}
        \centering
        \includegraphics[scale=0.2]{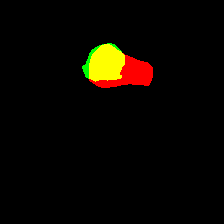}
    \end{minipage}
    \hspace{0.1em}
    \begin{minipage}[t]{0.08\linewidth}
        \centering
        \includegraphics[scale=0.2]{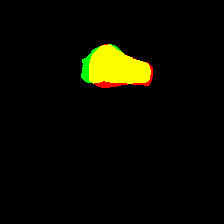}
    \end{minipage}
    \hspace{0.1em}
    \begin{minipage}[t]{0.08\linewidth}
        \centering
        \includegraphics[scale=0.2]{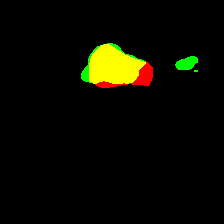}
    \end{minipage}
    \hspace{0.1em}
    \begin{minipage}[t]{0.08\linewidth}
        \centering
        \includegraphics[scale=0.2]{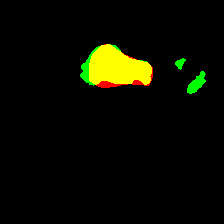}
    \end{minipage}
    \hspace{0.1em}
    \begin{minipage}[t]{0.08\linewidth}
        \centering
        \includegraphics[scale=0.2]{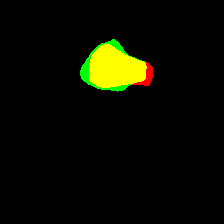}
    \end{minipage}
    \par   
    \begin{minipage}[t]{0.08\linewidth}   
        \centering
        \includegraphics[scale=0.2]{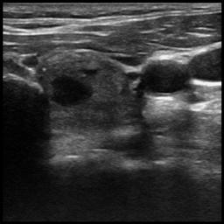}
    \end{minipage}
    \hspace{0.1em}
    \begin{minipage}[t]{0.08\linewidth}
        \centering
        \includegraphics[scale=0.2]{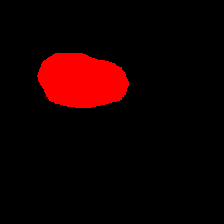}
    \end{minipage}
    \hspace{0.1em}
    \begin{minipage}[t]{0.08\linewidth}
        \centering
        \includegraphics[scale=0.2]{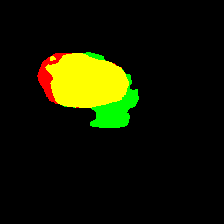}
    \end{minipage}
    \hspace{0.1em}
    \begin{minipage}[t]{0.08\linewidth}
        \centering
        \includegraphics[scale=0.2]{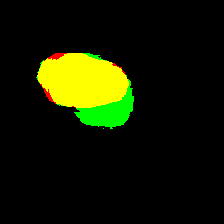}
    \end{minipage}
    \hspace{0.1em}
    \begin{minipage}[t]{0.08\linewidth}
        \centering
        \includegraphics[scale=0.2]{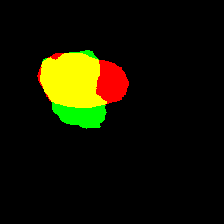}
    \end{minipage}
    \hspace{0.1em}
    \begin{minipage}[t]{0.08\linewidth}
        \centering
        \includegraphics[scale=0.2]{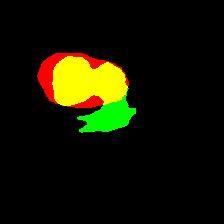}
    \end{minipage}
    \hspace{0.1em}
    \begin{minipage}[t]{0.08\linewidth}
        \centering
        \includegraphics[scale=0.2]{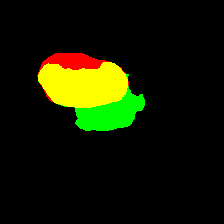}
    \end{minipage}
    \hspace{0.1em}
    \begin{minipage}[t]{0.08\linewidth}
        \centering
        \includegraphics[scale=0.2]{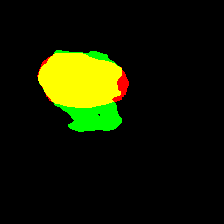}
    \end{minipage}
    \hspace{0.1em}
    \begin{minipage}[t]{0.08\linewidth}
        \centering
        \includegraphics[scale=0.2]{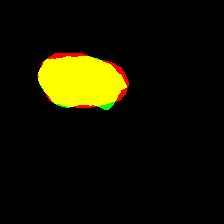}
    \end{minipage}
    \par  
\begin{minipage}{\linewidth}
    \centering
    \begin{minipage}[t]{0.08\linewidth}
        \centering
        \makebox[0.08\linewidth]{Image}
    \end{minipage}
    \hspace{0.1em}
    \begin{minipage}[t]{0.08\linewidth}
        \centering
        \makebox[0.08\linewidth]{GT}
    \end{minipage}
    \hspace{0.1em}
    \begin{minipage}[t]{0.08\linewidth}
        \centering
        \makebox[0.08\linewidth]{CPS}
    \end{minipage}
    \hspace{0.1em}
    \begin{minipage}[t]{0.08\linewidth}
        \centering
        \makebox[0.08\linewidth]{ABD}
    \end{minipage}
    \hspace{0.1em}
    \begin{minipage}[t]{0.08\linewidth}
        \centering
        \makebox[0.08\linewidth]{AD-MT}
    \end{minipage}
    \hspace{0.1em}
    \begin{minipage}[t]{0.08\linewidth}
        \centering
        \makebox[0.08\linewidth]{DiffRect}
    \end{minipage}
    \hspace{0.1em}
    \begin{minipage}[t]{0.08\linewidth}
        \centering
        \makebox[0.08\linewidth]{DCNet}
    \end{minipage}
    \hspace{0.1em}
    \begin{minipage}[t]{0.08\linewidth}
        \centering
        \makebox[0.08\linewidth]{SS-Net}
    \end{minipage}
    \hspace{0.1em}
    \begin{minipage}[t]{0.08\linewidth}
        \centering
        \makebox[0.08\linewidth]{\textbf{Hermes}}
    \end{minipage}
    \end{minipage}
    \vspace{0.5em}
    \caption{Visual comparison of various state-of-the-art methods on the BUSI, UDIAT and SZ-TUS datasets. The first two rows correspond to the BUSI dataset, where models are trained on 72 labeled images. The third and fourth rows show results from the UDIAT dataset, with models trained on 40 labeled images. The final row showcases results from the SZ-TUS dataset, with the model trained on 60 labeled images. In the visualizations, red, green, and yellow regions indicate the ground truth, predictions and overlapping areas, respectively.}
    \vspace{1.0em}
    \label{fig:4}
\end{figure*}

To intuitively highlight the superiority of Hermes, Fig.~\ref{fig:4} presents segmentation results across all compared methods. Hermes consistently achieves more accurate segmentations, regardless of variations in target scale or boundary clarity.
\\
\textbf{Classification result.} We compare Hermes with four recent semi-supervised classification approaches. ShrinkMatch \cite{bd:17} and InterLUDE \cite{bd:16} focus on enhancing representation learning to better capture underlying data structures, while PEFAT \cite{bd:14} integrates feature adversarial training with pseudo-loss estimation to improve model robustness. InfoMatch \cite{bd:15}, on the other hand, leverages information entropy estimation to optimize the use of unlabeled data.

The results in Tab. \ref{tab:2} clearly demonstrate Hermes' superior performance across all tested scenarios. On the BUSI dataset, Hermes consistently achieves the highest accuracy, with improvements ranging from 2.2\% to 4.45\% over the second-best method (InterLUDE\cite{bd:16}) as the number of labeled samples decreases from 216 to 144. This trend is even more pronounced on the SZ-TUS dataset, where Hermes outperforms other methods by a margin of 1.96\% to 5.63\% with 120 and 60 labels, respectively. Notably, in the most label-scarce scenario (SZ-TUS with 60 labels), Hermes' accuracy of 81.24\% represents a significant leap over competing methods, underscoring its ability to effectively leverage unlabeled data to compensate for the lack of supervision.


\begin{table*}[t]
\caption{Quantitative comparison of Dice rate (\%) across various state-of-the-art methods on the BUSI, UDIAT and SZ-TUS datasets.}
\vspace{0.5em}
\centering
\small
\begin{tabular}{c|lll|lll|ll}
\toprule
\multirow{2}{*}{Method} & \multicolumn{3}{c|}{BUSI} & \multicolumn{3}{c|}{UDIAT} & \multicolumn{2}{c}{SZ-TUS} \\ \cmidrule{2-9} 
 & \multicolumn{1}{c|}{216 labels} & \multicolumn{1}{c|}{144 labels} & \multicolumn{1}{c|}{72 labels} & \multicolumn{1}{c|}{60 labels} & \multicolumn{1}{c|}{40 labels} & \multicolumn{1}{c|}{20 labels} & \multicolumn{1}{c|}{120 labels} & \multicolumn{1}{c}{60 labels} \\ \midrule
SupOnly & \multicolumn{1}{l|}{83.62±0.21} & \multicolumn{1}{l|}{79.28±0.84} & 71.84±0.45 & \multicolumn{1}{l|}{85.78±0.39} & \multicolumn{1}{l|}{83.28±0.47} & 78.74±0.63 & \multicolumn{1}{l|}{77.24±0.59} & 73.42±0.41 \\ \midrule
CPS\cite{bd:8} & \multicolumn{1}{l|}{84.84±0.35} & \multicolumn{1}{l|}{80.21±0.48} & 72.21±0.61 & \multicolumn{1}{l|}{86.22±0.46} & \multicolumn{1}{l|}{84.03±0.48} & 79.52±0.38 & \multicolumn{1}{l|}{77.84±0.55} & 74.26±0.54 \\
ABD\cite{bd:9} & \multicolumn{1}{l|}{85.91±0.25} & \multicolumn{1}{l|}{82.77±0.56} & 73.93±0.44 & \multicolumn{1}{l|}{87.21±0.28} & \multicolumn{1}{l|}{85.57±0.35} & 81.73±0.43 & \multicolumn{1}{l|}{78.92±0.48} & 76.08±0.43 \\
AD-MT\cite{bd:12} & \multicolumn{1}{l|}{85.02±0.49} & \multicolumn{1}{l|}{81.27±0.61} & 72.64±0.32 & \multicolumn{1}{l|}{86.75±0.67} & \multicolumn{1}{l|}{84.51±0.56} & 80.54±0.55 & \multicolumn{1}{l|}{78.48±0.49} & 75.21±0.49 \\
DiffRect\cite{bd:11} & \multicolumn{1}{l|}{85.54±0.41} & \multicolumn{1}{l|}{82.41±0.53} & 73.48±0.59 & \multicolumn{1}{l|}{86.97±0.81} & \multicolumn{1}{l|}{85.12±0.41} & 81.21±0.44 & \multicolumn{1}{l|}{78.67±0.53} & 74.84±0.51 \\
DCNet\cite{bd:13} & \multicolumn{1}{l|}{85.11±0.63} & \multicolumn{1}{l|}{81.88±0.64} & 73.03±0.41 & \multicolumn{1}{l|}{87.52±0.53} & \multicolumn{1}{l|}{84.77±0.47} & 80.92±0.39 & \multicolumn{1}{l|}{78.06±0.55} & 74.47±0.52 \\
SS-Net\cite{bd:10} & \multicolumn{1}{l|}{85.27±0.52} & \multicolumn{1}{l|}{82.21±0.47} & 73.25±0.71 & \multicolumn{1}{l|}{86.42±0.62} & \multicolumn{1}{l|}{84.95±0.56} & 81.03±0.48 & \multicolumn{1}{l|}{78.29±0.51} & 75.13±0.47 \\ \midrule
\textbf{Hermes} & \multicolumn{1}{l|}{\textbf{87.48±0.47}} & \multicolumn{1}{l|}{\textbf{84.41±0.51}} & \textbf{75.68±0.22} & \multicolumn{1}{l|}{\textbf{88.73±0.48}} & \multicolumn{1}{l|}{\textbf{87.15±0.32}} & \textbf{83.23±0.28} & \multicolumn{1}{l|}{\textbf{80.23±0.46}} & \textbf{77.33±0.37} \\ \bottomrule
\end{tabular}
\label{tab:1}
\end{table*}

\begin{table}[t]
\caption{Quantitative comparison of accuracy rate (\%) across various state-of-the-art methods on the BUSI and SZ-TUS datasets.}
\vspace{0.3em}
\centering
\scriptsize
\begin{tabular}{c|cc|cc}
\toprule
\multirow{2}{*}{Method} & \multicolumn{2}{c|}{BUSI} & \multicolumn{2}{c}{SZ-TUS} \\ \cmidrule{2-5} 
 & \multicolumn{1}{c|}{216 labels} & 144 labels & \multicolumn{1}{c|}{120 labels} & 60 labels \\ \midrule
SupOnly & \multicolumn{1}{c|}{91.06±0.48} & 87.51±0.52 & \multicolumn{1}{c|}{78.96±0.31} & 75.61±0.39 \\ \midrule
PEFAT\cite{bd:14} & \multicolumn{1}{c|}{91.38±0.46} & 87.82±0.45 & \multicolumn{1}{c|}{79.67±0.42} & 76.27±0.53 \\
InfoMatch\cite{bd:15} & \multicolumn{1}{c|}{91.67±0.54} & 87.96±0.41 & \multicolumn{1}{c|}{80.15±0.46} & 77.97±0.47 \\
InterLUDE\cite{bd:16} & \multicolumn{1}{c|}{92.31±0.35} & 88.21±0.32 & \multicolumn{1}{c|}{81.69±0.36} & 79.32±0.37 \\
ShrinkMatch\cite{bd:17} & \multicolumn{1}{c|}{91.89±0.36} & 88.02±0.38 & \multicolumn{1}{c|}{80.89±0.47} & 78.48±0.44 \\ \midrule
\textbf{Hermes} & \multicolumn{1}{c|}{\textbf{94.51±0.32}} & \textbf{90.20±0.31} & \multicolumn{1}{c|}{\textbf{83.65±0.28}} & \textbf{81.24±0.29} \\ \bottomrule
\end{tabular}
\label{tab:2}
\end{table}
\begin{figure}[t]
    \centering
    \begin{subfigure}{0.45\linewidth}
        \centering
        \includegraphics[width=\linewidth]{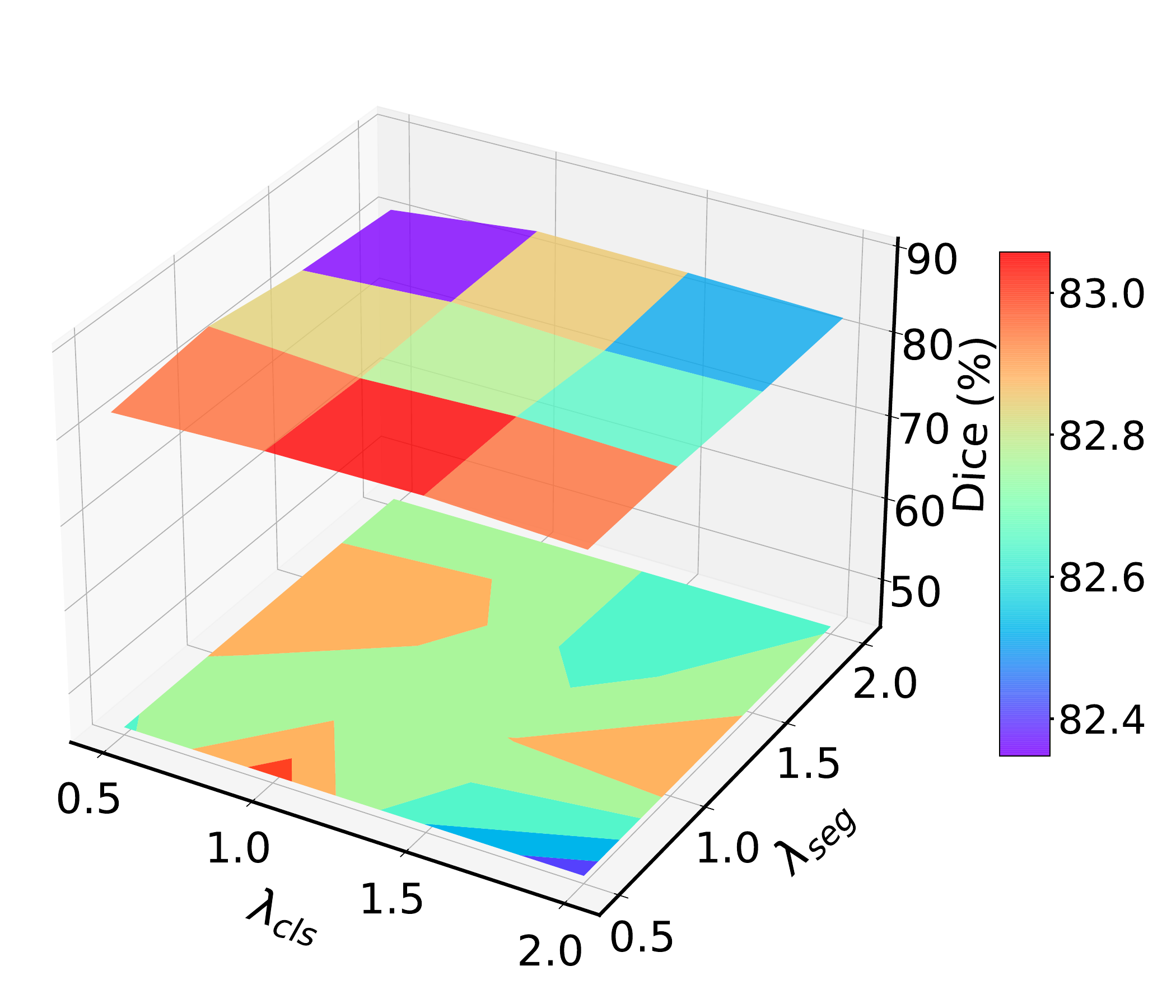} 
        \caption{Segmentation}
        \label{fig5:a}
    \end{subfigure}
    \hspace{0.05\linewidth} 
    \begin{subfigure}{0.45\linewidth}
        \centering
        \includegraphics[width=\linewidth]{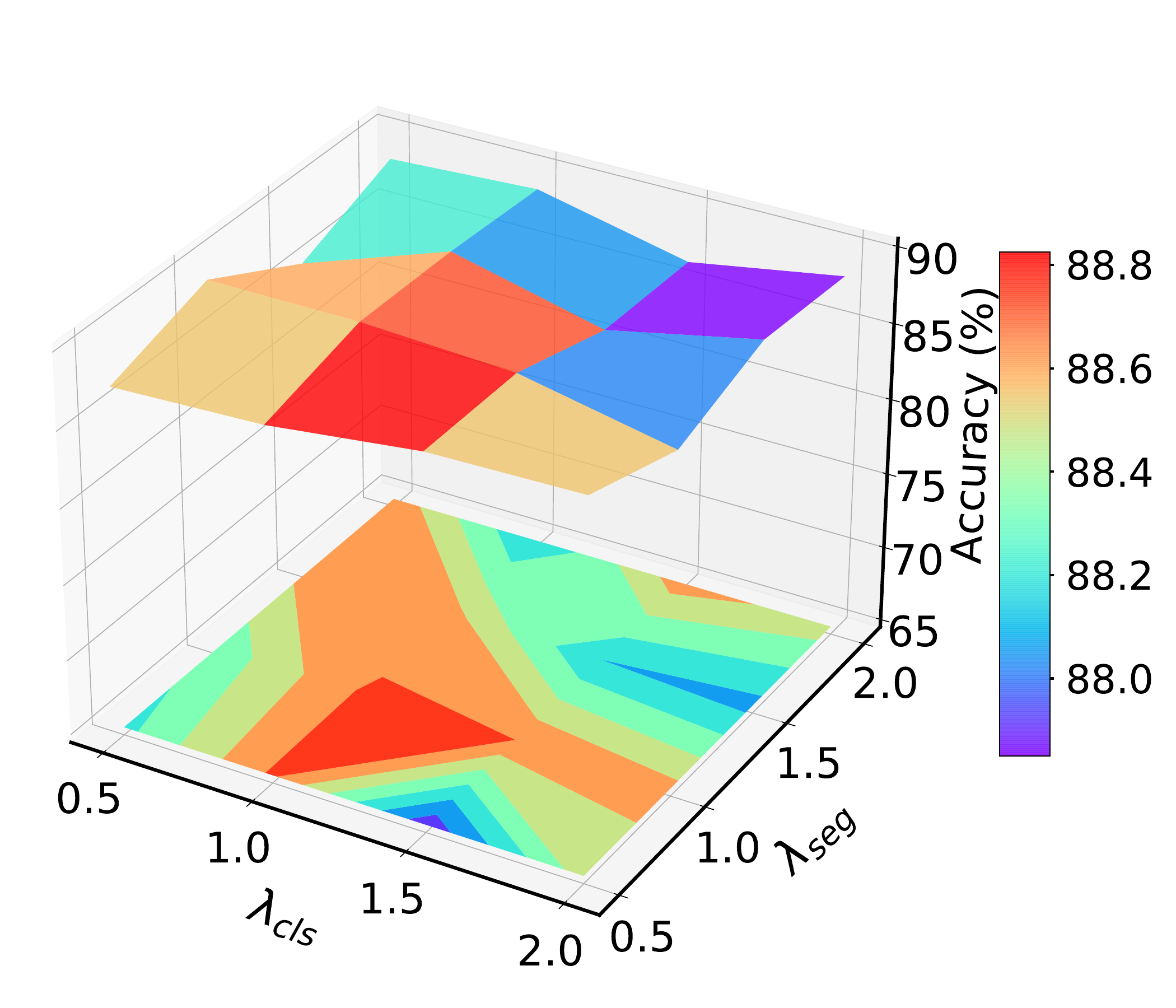} 
        \caption{Classification}
        \label{fig5:b}
    \end{subfigure}
    \vspace{1.3em}
    \caption{Sensitivity analysis of Hermes to hyper-parameters $\lambda_{cls}$ and $\lambda_{seg}$. All results are obtained using the BUSI dataset with 144 labeled images.}
    \label{fig:5}
    \vspace{2.0em}
\end{figure}
\subsection{Ablation Study}
\begin{table}[t]
\vspace{0.5em}
\caption{Ablation study of various components of segmentation and classification branch on the BUSI dataset using 72 labeled images. DTCL: Dual-Threshold Contrastive Learning. SM: Saliency Module. ITA: Inter-Task Attention. ITCL: Inter-Task Consistency Learning. SS: Single Segmentation. SC: Single Classification}
\vspace{0.5em}
\centering
\begin{tabular}{c|c|c|c|c|c|cc}
\toprule
\multirow{2}{*}{Index} & \multirow{2}{*}{DTCL} & \multirow{2}{*}{SM/ITA} & \multirow{2}{*}{ITCL} & SS & SC & \multicolumn{2}{c}{Joint Task} \\ \cmidrule{5-8} 
 &  &  &  & Dice & Acc & \multicolumn{1}{c|}{Dice} & Acc \\ \midrule
1 &  &  &  & 71.79 & 81.08 & \multicolumn{1}{c|}{71.84} & 81.17 \\
2 & \checkmark &  &  & 73.09 & 82.27 & \multicolumn{1}{c|}{73.13} & 82.32 \\
3 &  & \checkmark &  & - & - & \multicolumn{1}{c|}{72.95} & 82.39 \\
4 &  & \checkmark & \checkmark & - & - & \multicolumn{1}{c|}{74.09} & 83.43 \\
5 & \checkmark & \checkmark &  & - & - & \multicolumn{1}{c|}{74.24} & 83.54 \\
6 & \checkmark & \checkmark & \checkmark & - & - & \multicolumn{1}{c|}{\textbf{75.68}} & \textbf{84.77} \\ \bottomrule
\end{tabular}
\label{tab:ablation}
\vspace{0.4em}
\end{table}

\begin{figure}[t]
  \centering
  \includegraphics[width=\linewidth]{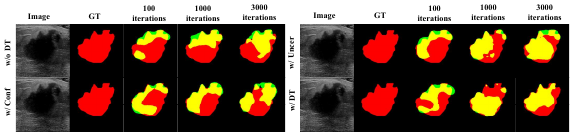}
    \vspace{0.01em}
   \caption{Visualization of different strategy on UDIAT dataset with 40 labeled images. w/o DT: without dual-threshold. w/ Conf: with confidence. w/ Uncer: with uncertainty. w/ DT: with dual-threshold.}
   \label{fig:contrast}
   \vspace{1.2em}
\end{figure}

\begin{figure}[t]
    \centering
    \begin{subfigure}{0.28\linewidth}
        \centering
        \includegraphics[width=0.70\linewidth]{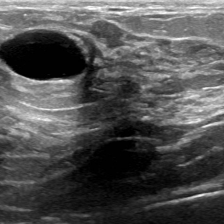} 
        \captionsetup{font={footnotesize}}
        \caption{Input image}
        \label{attenion:original}
    \end{subfigure}
        \begin{subfigure}{0.28\linewidth}
        \centering
        \includegraphics[width=0.70\linewidth]{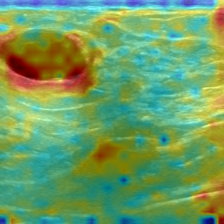} 
        \captionsetup{font={footnotesize}}
        \caption{Channel attention}
        \label{attention:spatial}
    \end{subfigure}
    \begin{subfigure}{0.28\linewidth}
        \centering
        \includegraphics[width=0.70\linewidth]{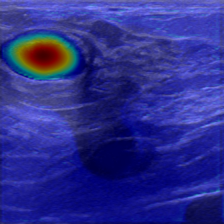} 
        \captionsetup{font={footnotesize}}
        \caption{Spatial attention}
        \label{attention:channel}
    \end{subfigure}
    \vspace{1.5em}
    \caption{Visualization results of the inter-task attention. The colored regions visualize the attention map, with a red-to-yellow gradient indicating a gradual decrease in attention weights.}
    \label{fig:attention}
    \vspace{2.0em}
\end{figure}

\textbf{Effectiveness of components.} Saliency module (SM) is used to segmentation branch, while Inter-task attention (ITA) is used to classification branch. Single classification (SC) represents individual classification branches measured in accuracy (Acc), while single segmentation (SS) denotes individual segmentation branches measured in dice coefficient (Dice). The first and second rows of Tab. \ref{tab:ablation} show that the performance of joint task is similar to that of SS and SC. We argue that this is due to the independent backbone, where without SM/ITA or inter-task consistency learning (ITCL), joint task performance equals that of independent tasks. The second and fifth rows of Tab. \ref{tab:ablation} show that the Dice and Acc of joint task outperform Dice of SS and Acc of SC. This indicates that SM/ITA of joint task efficiently provide complementary semantic information. SM/ITA and dual-threshold contrastive learning (DTCL) are individual modules, while ITCL is not. ITCL is designed to restrain features in latent space to better share information about tumor characteristics, such as shape, edge, and size (where SM/ITA is utilized). As the segmentation task needs to identify the spatial location and shape of the tumor, while the classification task requires determining whether the tumor is malignant or benign based on its morphology and characteristics. So informations sharing between two tasks is very important. This viewpoint can be verified by the third and fourth rows of Tab. \ref{tab:ablation}.
\\
\textbf{Parameter sensitivity.} Fig. \ref{fig5:a} and Fig. \ref{fig5:b} show the sensitivity of Hermes to hyper-parameters $\lambda_{cls}$ and $\lambda_{seg}$, with values ranging from \{0.5, 1.0, 1.5, 2.0\}. The results indicate that Hermes maintains robust accuracy across a wide range of hyper-parameter values, specifically when $1.0 \leq \lambda_{cls} \leq 1.5$ and $0.5 \leq \lambda_{seg} \leq 1.0$. The optimal segmentation performance is achieved when $\lambda_{cls}=1.0$ and $\lambda_{seg}=0.5$.
\\
\textbf{Effectiveness of dual-threshold selection.} As shown in Figure~\ref{fig:contrast}, red, green, and yellow regions denote ground truth, false predictions, and correct predictions, respectively. Dual-threshold selection yields cleaner pseudo-masks by reducing inaccurate labels. This is because high-confidence predictions may still be unreliable in early training. To improve pseudo-label quality, we introduce uncertainty as an additional constraint and retain only those with both high confidence and low uncertainty.
\\
\textbf{Visualization of inter-task attention.} To better demonstrate the advantages of inter-task attention, Fig. \ref{fig:attention} illustrates visualizations of the attention map of channel attention in Eq. (\ref{eq:3}) and spatial attention in Eq. (\ref{eq:4}). Heat map is used to represent the attention map. The channel attention map assigns high weights to both the tumor and surrounding regions, reflecting its focus on global feature contributions. In contrast, the spatial attention map is more localized to the tumor, with concentrated high-weight (red) regions and sharper boundaries, as it directly captures spatial information. Thus, the visualization confirms the effectiveness of inter-task attention, as it successfully highlights the tumor region.

More hyper-parameters analysis and ablation studies can be found in supplementary material.
\section{Conclusion}
It is known that confidence-based pseudo-label selection often produces overly confident yet inaccurate predictions, degrading the performance of semi-supervised contrastive learning. Moreover, the discriminative information between segmentation and classification tasks remains underexplored. To address these challenges, we propose a novel semi-supervised dual-threshold contrastive learning framework for ultrasound image classification and segmentation. An inter-task attention and saliency module is designed to promote information sharing between tasks, while an inter-task consistency strategy is introduced to align tumor features and mitigate negative transfer. In addition, we release a new thyroid ultrasound dataset, SZ-TUS, to support future research. Extensive experiments on internal and external datasets demonstrate that Hermes achieves state-of-the-art performance and validate the effectiveness of our joint learning strategy.
\begin{ack}
This work was supported in part by the Natural Science Foundation of China under Grant 62476175 and Grant 62272319, and in part by the Natural Science Foundation of Guangdong Province (Grant 2023A1515010677, 2024A1515011637, 2023B1212060076), the Science and Technology Planning Project of Shenzhen Municipality under Grant JCYJ20220818095803007, and the Special Project for Clinical and Basic Sci\&Tech Innovation of Guangdong Medical University under Grant GDMULCJC2024151.
\end{ack}

\bibliography{mybibfile}
\newpage

\twocolumn[
\begin{center}
    \Huge\bfseries Supplementary Material \par
\end{center}
]

\section{More hyper-parameters analysis}
\begin{figure}[ht]
    \centering
    \includegraphics[width=\columnwidth]{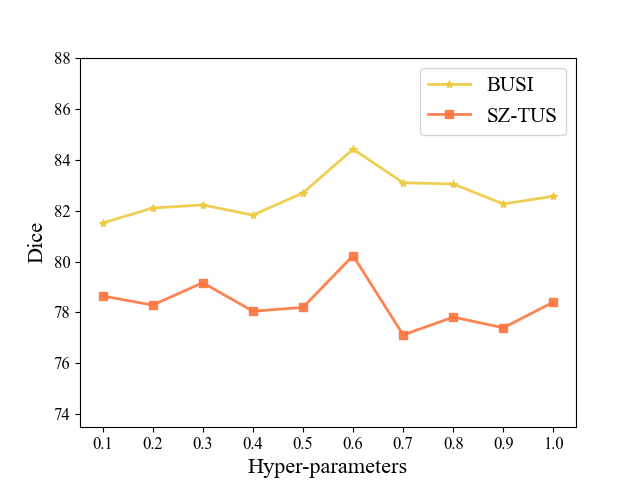}
    \vspace{0.05em}
    \caption{Ablation study of pixel-level contrastive weight on the BUSI dataset with 144 labeled images and the SZ-TUS dataset with 120 labeled images.}
    \label{sup:fig1}
    \vspace{1.5em}
\end{figure}

\begin{figure}[ht]
    \centering
    \includegraphics[width=\columnwidth]{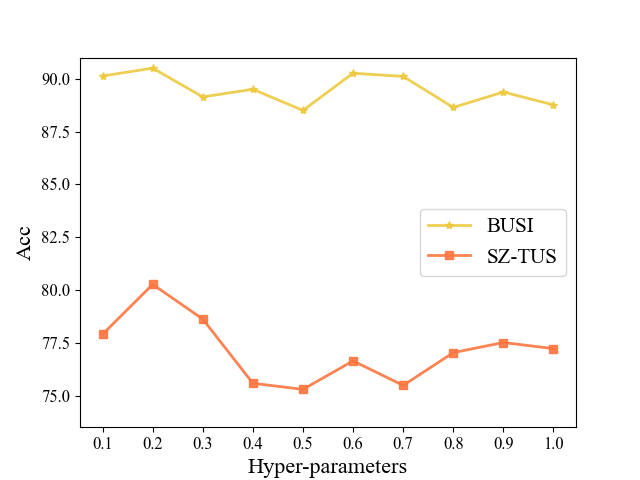}
    \vspace{0.05em}
    \caption{Ablation study of image-level contrastive weight on the BUSI dataset with 144 labeled images and the SZ-TUS dataset with 120 labeled images.}
    \label{sup:fig2}
    \vspace{2.0em}
\end{figure}

\begin{figure}[ht]
    \centering
    \includegraphics[width=\columnwidth]{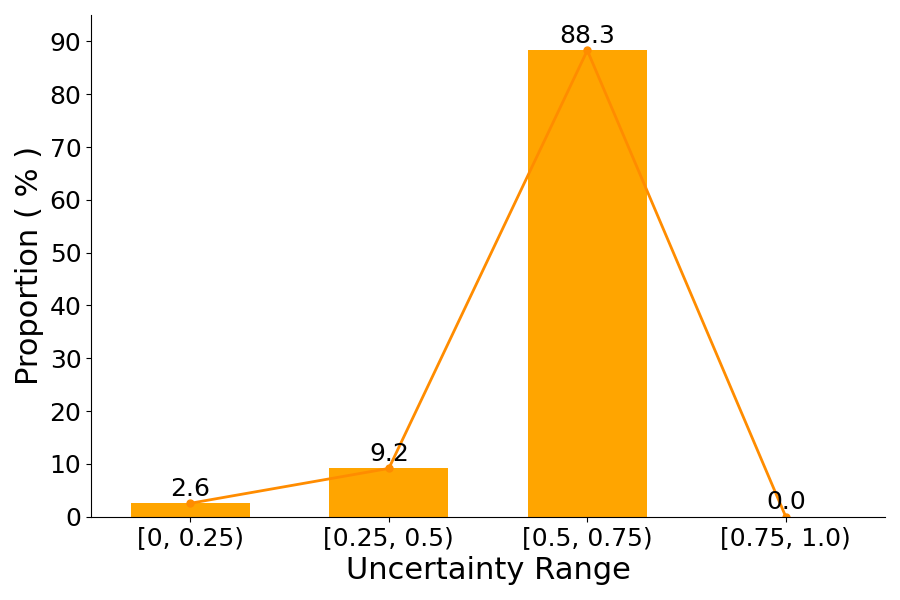}
    \vspace{0.05em}
    \caption{The uncertainty distribution of prediction in UDIAT dataset.}
    \label{sup:uncertainty}
    \vspace{2.0em}
\end{figure}

\textbf{Contrastive learning weight.} The weight assigned to contrastive learning loss is crucial for optimizing our method's performance. A higher contrastive learning loss weight strengthens the model's feature representation and robustness but may hinder the performance of supervised tasks. Conversely, a lower contrastive loss weight shifts the model’s reliance towards supervised signals, potentially missing out on valuable information from unlabeled data and the feature space. Obtaining a balanced weight is important to harmonizing both classification and segmentation tasks, allowing the model to recognize different classes while maintaining precise segmentation of intricate details and boundaries. Fig. \ref{sup:fig1} and Fig. \ref{sup:fig2} illustrate the ablation study results for different values of $\alpha$ and $\beta$. The optimal performance is achieved when $\alpha$ is set to 0.6 and $\beta$ to 0.2, demonstrating the significance of appropriately tuning these hyperparameters for our method.
\begin{table}[h]
\centering
\vspace{1.0em}
\caption{Ablation study of learning rate on the BUSI dataset with 72 labeled images.}
\vspace{1.0em}
\begin{tabular}{c|ccccc}
\toprule
LR & 8e-5 & 9e-5 & \textbf{1e-4} & 2e-4 & 3e-4 \\ \midrule
Dice(\%) & 73.91 & 74.35 & \textbf{75.68} & 73.93 & 73.22 \\
Acc(\%) & 81.86 & 82.17 & \textbf{83.19} & 82.76 & 82.50 \\ \bottomrule
\end{tabular}
\label{sup:tab1}
\end{table}
\\
\textbf{Learning rate.} In image segmentation and classification tasks, the learning rate plays a crucial role in influencing the model's training speed, convergence, performance, and generalization. A higher learning rate can expedite training but may result in instability and overlook finer details. Conversely, a lower learning rate promotes more stable training and captures intricate features, though it may get trapped in local optima. Achieving the optimal balance is crucial for enhancing both performance and training efficiency. We conduct an ablation study on different learning rates (LR) in Tab. \ref{sup:tab1}, demonstrating that our method yields optimal performance when the learning rate for segmentation and classification branches is set to 1e-4.
\\
\textbf{Uncertainty threshold.} We analyze the uncertainty distribution during the first epoch of training on the UDIAT dataset. The uncertainty histogram, shown in Fig. \ref{sup:uncertainty}, is computed using Eq. (8) from the main text. The results indicate that all uncertainty scores fall below 0.75. Accordingly, we set the initial uncertainty threshold to 0.75, which is dynamically adjusted throughout training. The formulation is as follows:
\begin{equation}
    \kappa =\eta_{min} +0.5\times \left ( \eta_{max}-\eta_{min} \right )  \times \left [cosine\left ( \frac{iter}{iter_{total}}\cdot \pi   \right ) +1  \right ] 
\end{equation}
where $\eta_{max}$ and $\eta_{min}$ are set to 0.75 and 0.25, respectively. $iter$ is the current number of iterations, while $iter_{all}$ is the total number of iterations.
\begin{figure}[ht]
    \centering
    \includegraphics[width=\columnwidth]{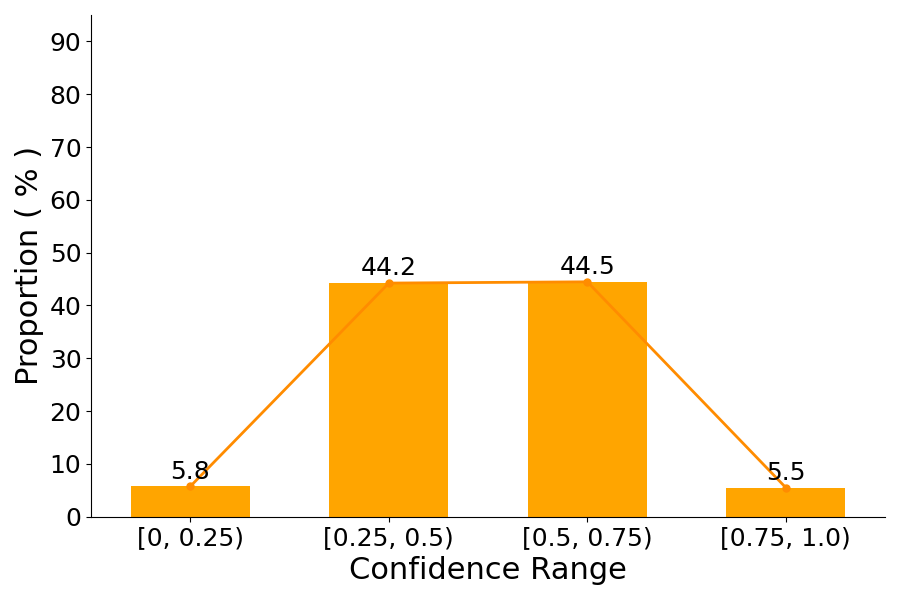}
    \vspace{0.05em}
    \caption{The confidence distribution of prediction in UDIAT dataset.}
    \label{sup:confidence}
    \vspace{2.0em}
\end{figure}
\\
\textbf{Confidence threshold.} We also analyze the confidence distribution during the first epoch of training on the UDIAT dataset. The confidence histogram, shown in Fig. \ref{sup:confidence}. Although the most confidence scores are below 0.75, we set the initial confidence threshold to 0.75. A small confidence threshold results in incorrect pseudo-label learning for the model, while too high a threshold makes it difficult for the model to acquire learning information. Thus, dynamic confidence threshold can effectively improve the performance of model from learning unlabeled data. The formulation is as follows:
\begin{equation}
    \tau=0.75+0.25\times e^{-0.5\times \left ( 1-\frac{iter}{iter_{all}}  \right )^{2} }\times ln2
\end{equation}
where ln2 is the constant. $iter$ is the current number of iterations, while $iter_{all}$ is the total number of iterations.


\section{More ablation studies}
\textbf{Segmentation contrast on different feature scales.} The resolution of feature maps plays a crucial role in selecting samples for contrastive learning. To identify the optimal feature scale, we evaluate the effect of different feature resolutions on contrastive learning using the BUSI dataset. As shown in Tab. \ref{sup:tab2}, contrastive learning under low-resolution features (Conv4) yields inferior performance, likely due to semantic inconsistencies introduced by label downsampling. On the other hand, using high-resolution features (Conv1) also results in degraded performance, possibly because high-resolution pixel vectors lack sufficient semantic information. Therefore, our findings suggest that employing mid-level features for contrastive learning offers superior segmentation performance.
\begin{table}
\centering
\caption{Comparison of pixel contrastive learning performance across different feature resolutions on the BUSI dataset with 144 labeled images.}
\vspace{1.0em}
\begin{tabular}{c|c|cc}
\toprule
\multirow{2}{*}{Feature} & \multirow{2}{*}{Resolution} & \multicolumn{2}{c}{Metrics} \\ \cmidrule{3-4} 
 &  & \multicolumn{1}{c|}{Dice(\%)} & IoU(\%) \\ \midrule
$Conv_{1}$ & $56\times56$ & \multicolumn{1}{c|}{81.26} & 69.22 \\
$Conv_{2}$ & $28\times28$ & \multicolumn{1}{c|}{81.56} & 69.62 \\
\textbf{$\bm{Conv_{3}}$} & \bm{$14\times14$} & \multicolumn{1}{c|}{\textbf{83.37}} & \textbf{72.54} \\
$Conv_{4}$ & $7\times7$ & \multicolumn{1}{c|}{80.72} & 68.06 \\ \bottomrule
\end{tabular}
\vspace{1.0em}
\label{sup:tab2}
\end{table}
\begin{table}
\centering
\caption{Comparison of saliency map merging with different feature resolutions on the BUSI dataset using 144 labeled images.}
\vspace{1.0em}
\begin{tabular}{c|c|cc}
\toprule
\multirow{2}{*}{Feature} & \multirow{2}{*}{Resolution} & \multicolumn{2}{c}{Metrics} \\ \cmidrule{3-4} 
 &  & \multicolumn{1}{c|}{Dice(\%)} & IoU(\%) \\ \midrule
$Conv\_up_{1}$ & $14\times14$ & \multicolumn{1}{c|}{82.28} & 70.04 \\
$Conv\_up_{2}$ & $28\times28$ & \multicolumn{1}{c|}{81.46} & 69.25 \\
\textbf{$\bm{Conv\_up_{3}}$} & \textbf{$ \bm{56\times56}$} & \multicolumn{1}{c|}{\textbf{83.14}} & \textbf{71.56} \\
$Conv\_up_{4}$ & $112\times112$ & \multicolumn{1}{c|}{82.53} & 70.61 \\ \bottomrule
\end{tabular}
\label{sup:tab3}
\end{table}
\\
\textbf{Saliency map fusion on different feature scales.} The saliency map generated by the classification branch plays a critical role in merging features from different resolutions in the decoder of the segmentation branch. To determine the optimal feature scale for fusion, we conduct an experiment analyzing the effect of merging features at various stages of the decoder. The results, presented in Tab. \ref{sup:tab3}, reveal that the best fusion occurs in the mid-to-late layers of the segmentation branch’s decoder. Thses fusion points are ideal as they strike a balance between the high-level semantic information from the classification branch and the detailed spatial information recovered by the segmentation branch. By integrating features at this stage, the model is able to leverage both the fine-grained details necessary for accurate segmentation and the semantic context needed for better classification performance. This finding underscores the importance of effective multi-resolution fusion in enhancing the synergy between segmentation and classification tasks.

\section{More visual comparison results}
\begin{figure*}
    \centering
    \begin{minipage}[t]{0.08\linewidth}   
        \centering
        \includegraphics[scale=0.2]{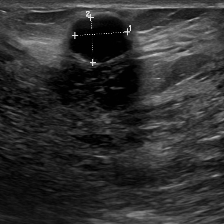}
    \end{minipage}
    \hspace{0.2em}
    \begin{minipage}[t]{0.08\linewidth}
        \centering
        \includegraphics[scale=0.2]{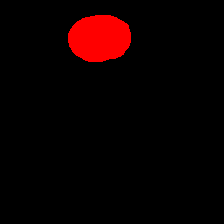}
    \end{minipage}
    \hspace{0.2em}
    \begin{minipage}[t]{0.08\linewidth}
        \centering
        \includegraphics[scale=0.2]{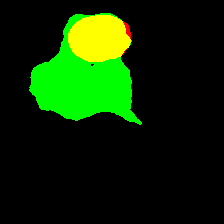}
    \end{minipage}
    \hspace{0.2em}
    \begin{minipage}[t]{0.08\linewidth}
        \centering
        \includegraphics[scale=0.2]{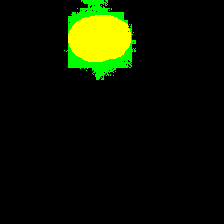}
    \end{minipage}
    \hspace{0.2em}
    \begin{minipage}[t]{0.08\linewidth}
        \centering
        \includegraphics[scale=0.2]{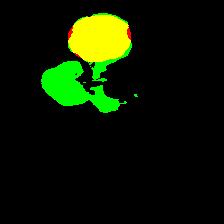}
    \end{minipage}
    \hspace{0.2em}
    \begin{minipage}[t]{0.08\linewidth}
        \centering
        \includegraphics[scale=0.2]{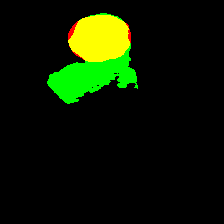}
    \end{minipage}
    \hspace{0.2em}
    \begin{minipage}[t]{0.08\linewidth}
        \centering
        \includegraphics[scale=0.2]{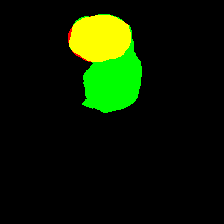}
    \end{minipage}
    \hspace{0.2em}
    \begin{minipage}[t]{0.08\linewidth}
        \centering
        \includegraphics[scale=0.2]{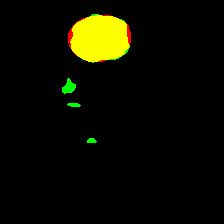}
    \end{minipage}
    \hspace{0.2em}
    \begin{minipage}[t]{0.08\linewidth}
        \centering
        \includegraphics[scale=0.2]{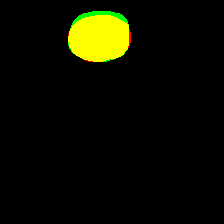}
    \end{minipage}
    \par   
    \vspace{0.1em}
    \begin{minipage}[t]{0.08\linewidth}   
        \centering
        \includegraphics[scale=0.2]{supple_figure/BUSI/original_image/val_79_input.png}
    \end{minipage}
    \hspace{0.2em}
    \begin{minipage}[t]{0.08\linewidth}
        \centering
        \includegraphics[scale=0.2]{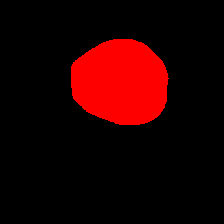}
    \end{minipage}
    \hspace{0.2em}
    \begin{minipage}[t]{0.08\linewidth}
        \centering
        \includegraphics[scale=0.2]{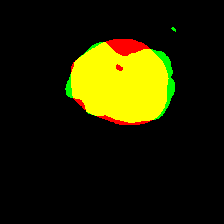}
    \end{minipage}
    \hspace{0.2em}
    \begin{minipage}[t]{0.08\linewidth}
        \centering
        \includegraphics[scale=0.2]{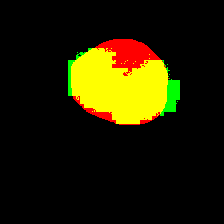}
    \end{minipage}
    \hspace{0.2em}
    \begin{minipage}[t]{0.08\linewidth}
        \centering
        \includegraphics[scale=0.2]{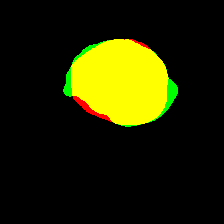}
    \end{minipage}
    \hspace{0.2em}
    \begin{minipage}[t]{0.08\linewidth}
        \centering
        \includegraphics[scale=0.2]{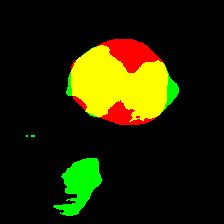}
    \end{minipage}
    \hspace{0.2em}
    \begin{minipage}[t]{0.08\linewidth}
        \centering
        \includegraphics[scale=0.2]{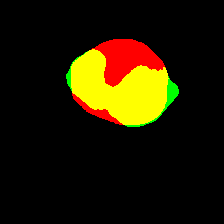}
    \end{minipage}
    \hspace{0.2em}
    \begin{minipage}[t]{0.08\linewidth}
        \centering
        \includegraphics[scale=0.2]{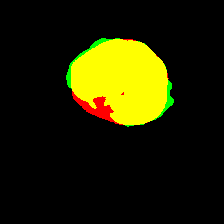}
    \end{minipage}
    \hspace{0.2em}
    \begin{minipage}[t]{0.08\linewidth}
        \centering
        \includegraphics[scale=0.2]{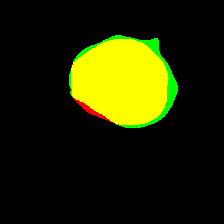}
    \end{minipage}
    \par   
    \vspace{0.1em}
    \begin{minipage}[t]{0.08\linewidth}   
        \centering
        \includegraphics[scale=0.2]{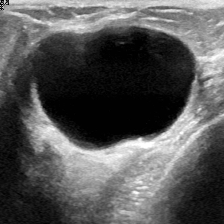}
    \end{minipage}
    \hspace{0.2em}
    \begin{minipage}[t]{0.08\linewidth}
        \centering
        \includegraphics[scale=0.2]{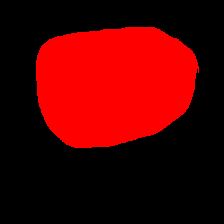}
    \end{minipage}
    \hspace{0.2em}
    \begin{minipage}[t]{0.08\linewidth}
        \centering
        \includegraphics[scale=0.2]{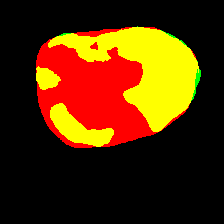}
    \end{minipage}
    \hspace{0.2em}
    \begin{minipage}[t]{0.08\linewidth}
        \centering
        \includegraphics[scale=0.2]{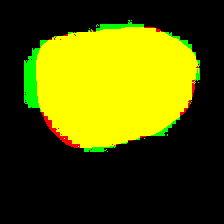}
    \end{minipage}
    \hspace{0.2em}
    \begin{minipage}[t]{0.08\linewidth}
        \centering
        \includegraphics[scale=0.2]{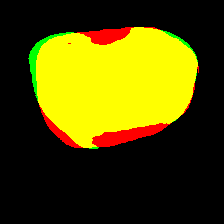}
    \end{minipage}
    \hspace{0.2em}
    \begin{minipage}[t]{0.08\linewidth}
        \centering
        \includegraphics[scale=0.2]{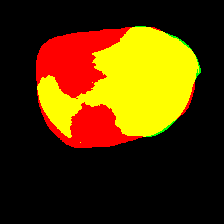}
    \end{minipage}
    \hspace{0.2em}
    \begin{minipage}[t]{0.08\linewidth}
        \centering
        \includegraphics[scale=0.2]{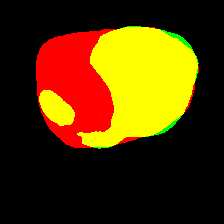}
    \end{minipage}
    \hspace{0.2em}
    \begin{minipage}[t]{0.08\linewidth}
        \centering
        \includegraphics[scale=0.2]{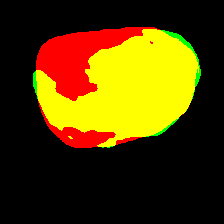}
    \end{minipage}
    \hspace{0.2em}
    \begin{minipage}[t]{0.08\linewidth}
        \centering
        \includegraphics[scale=0.2]{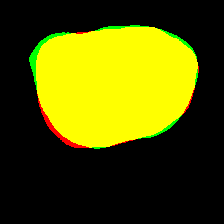}
    \end{minipage}
    \par   
    \vspace{0.1em}
    \begin{minipage}[t]{0.08\linewidth}   
        \centering
        \includegraphics[scale=0.2]{supple_figure/BUSI/original_image/val_175_input.png}
    \end{minipage}
    \hspace{0.2em}
    \begin{minipage}[t]{0.08\linewidth}
        \centering
        \includegraphics[scale=0.2]{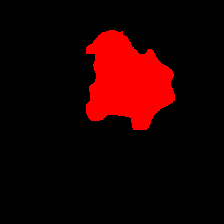}
    \end{minipage}
    \hspace{0.2em}
    \begin{minipage}[t]{0.08\linewidth}
        \centering
        \includegraphics[scale=0.2]{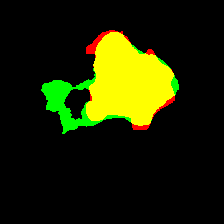}
    \end{minipage}
    \hspace{0.2em}
    \begin{minipage}[t]{0.08\linewidth}
        \centering
        \includegraphics[scale=0.2]{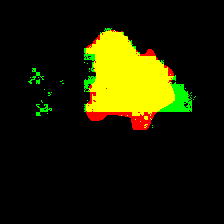}
    \end{minipage}
    \hspace{0.2em}
    \begin{minipage}[t]{0.08\linewidth}
        \centering
        \includegraphics[scale=0.2]{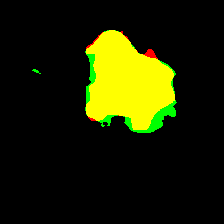}
    \end{minipage}
    \hspace{0.2em}
    \begin{minipage}[t]{0.08\linewidth}
        \centering
        \includegraphics[scale=0.2]{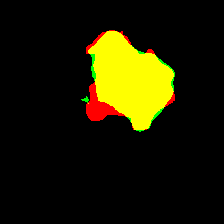}
    \end{minipage}
    \hspace{0.2em}
    \begin{minipage}[t]{0.08\linewidth}
        \centering
        \includegraphics[scale=0.2]{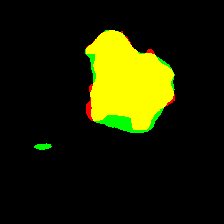}
    \end{minipage}
    \hspace{0.2em}
    \begin{minipage}[t]{0.08\linewidth}
        \centering
        \includegraphics[scale=0.2]{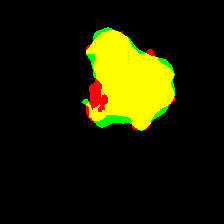}
    \end{minipage}
    \hspace{0.2em}
    \begin{minipage}[t]{0.08\linewidth}
        \centering
        \includegraphics[scale=0.2]{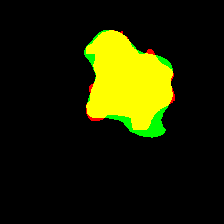}
    \end{minipage}
    \par   
    \vspace{0.1em}
    \begin{minipage}[t]{0.08\linewidth}   
        \centering
        \includegraphics[scale=0.2]{supple_figure/BUSI/original_image/val_125_input.png}
    \end{minipage}
    \hspace{0.2em}
    \begin{minipage}[t]{0.08\linewidth}
        \centering
        \includegraphics[scale=0.2]{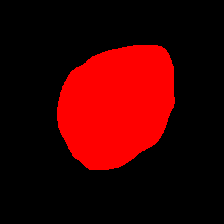}
    \end{minipage}
    \hspace{0.2em}
    \begin{minipage}[t]{0.08\linewidth}
        \centering
        \includegraphics[scale=0.2]{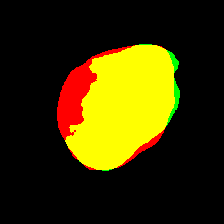}
    \end{minipage}
    \hspace{0.2em}
    \begin{minipage}[t]{0.08\linewidth}
        \centering
        \includegraphics[scale=0.2]{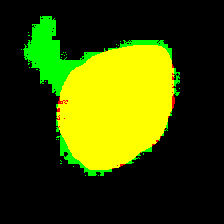}
    \end{minipage}
    \hspace{0.2em}
    \begin{minipage}[t]{0.08\linewidth}
        \centering
        \includegraphics[scale=0.2]{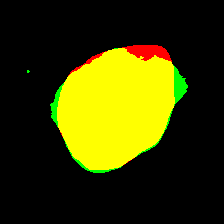}
    \end{minipage}
    \hspace{0.2em}
    \begin{minipage}[t]{0.08\linewidth}
        \centering
        \includegraphics[scale=0.2]{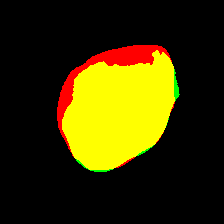}
    \end{minipage}
    \hspace{0.2em}
    \begin{minipage}[t]{0.08\linewidth}
        \centering
        \includegraphics[scale=0.2]{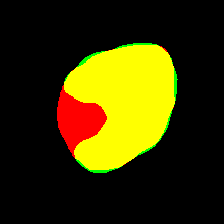}
    \end{minipage}
    \hspace{0.2em}
    \begin{minipage}[t]{0.08\linewidth}
        \centering
        \includegraphics[scale=0.2]{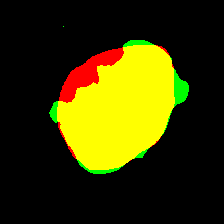}
    \end{minipage}
    \hspace{0.2em}
    \begin{minipage}[t]{0.08\linewidth}
        \centering
        \includegraphics[scale=0.2]{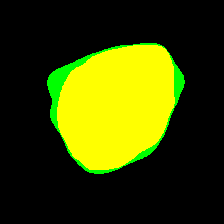}
    \end{minipage}
    \par   
    \vspace{0.1em}
    \begin{minipage}[t]{0.08\linewidth}   
        \centering
        \includegraphics[scale=0.2]{supple_figure/BUSI/original_image/val_50_input.png}
    \end{minipage}
    \hspace{0.2em}
    \begin{minipage}[t]{0.08\linewidth}
        \centering
        \includegraphics[scale=0.2]{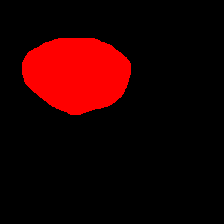}
    \end{minipage}
    \hspace{0.2em}
    \begin{minipage}[t]{0.08\linewidth}
        \centering
        \includegraphics[scale=0.2]{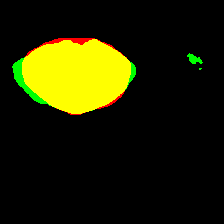}
    \end{minipage}
    \hspace{0.2em}
    \begin{minipage}[t]{0.08\linewidth}
        \centering
        \includegraphics[scale=0.2]{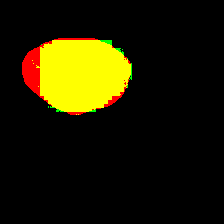}
    \end{minipage}
    \hspace{0.2em}
    \begin{minipage}[t]{0.08\linewidth}
        \centering
        \includegraphics[scale=0.2]{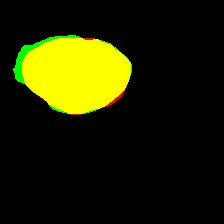}
    \end{minipage}
    \hspace{0.2em}
    \begin{minipage}[t]{0.08\linewidth}
        \centering
        \includegraphics[scale=0.2]{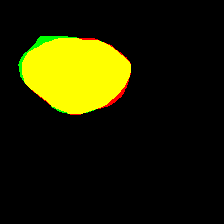}
    \end{minipage}
    \hspace{0.2em}
    \begin{minipage}[t]{0.08\linewidth}
        \centering
        \includegraphics[scale=0.2]{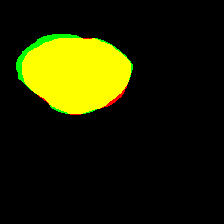}
    \end{minipage}
    \hspace{0.2em}
    \begin{minipage}[t]{0.08\linewidth}
        \centering
        \includegraphics[scale=0.2]{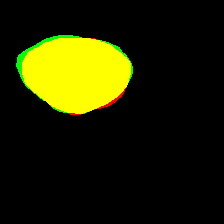}
    \end{minipage}
    \hspace{0.2em}
    \begin{minipage}[t]{0.08\linewidth}
        \centering
        \includegraphics[scale=0.2]{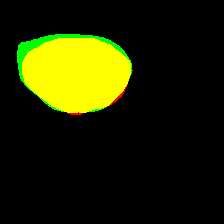}
    \end{minipage}
    \par   
    \vspace{0.1em}
    \begin{minipage}[t]{0.08\linewidth}   
        \centering
        \includegraphics[scale=0.2]{supple_figure/BUSI/original_image/val_54_input.png}
    \end{minipage}
    \hspace{0.2em}
    \begin{minipage}[t]{0.08\linewidth}
        \centering
        \includegraphics[scale=0.2]{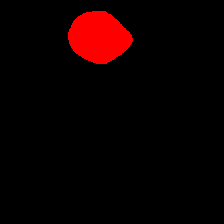}
    \end{minipage}
    \hspace{0.2em}
    \begin{minipage}[t]{0.08\linewidth}
        \centering
        \includegraphics[scale=0.2]{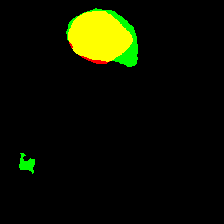}
    \end{minipage}
    \hspace{0.2em}
    \begin{minipage}[t]{0.08\linewidth}
        \centering
        \includegraphics[scale=0.2]{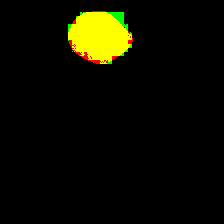}
    \end{minipage}
    \hspace{0.2em}
    \begin{minipage}[t]{0.08\linewidth}
        \centering
        \includegraphics[scale=0.2]{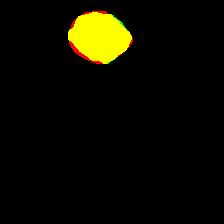}
    \end{minipage}
    \hspace{0.2em}
    \begin{minipage}[t]{0.08\linewidth}
        \centering
        \includegraphics[scale=0.2]{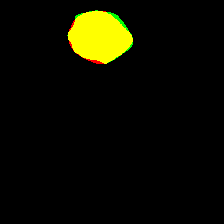}
    \end{minipage}
    \hspace{0.2em}
    \begin{minipage}[t]{0.08\linewidth}
        \centering
        \includegraphics[scale=0.2]{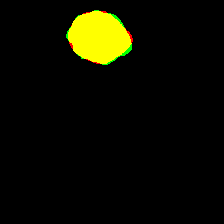}
    \end{minipage}
    \hspace{0.2em}
    \begin{minipage}[t]{0.08\linewidth}
        \centering
        \includegraphics[scale=0.2]{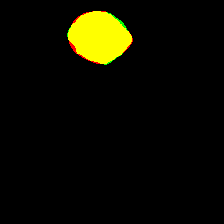}
    \end{minipage}
    \hspace{0.2em}
    \begin{minipage}[t]{0.08\linewidth}
        \centering
        \includegraphics[scale=0.2]{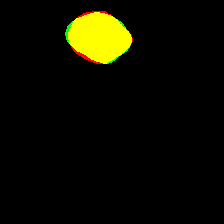}
    \end{minipage}
    \par   
    \vspace{0.1em}
    \begin{minipage}[t]{0.08\linewidth}   
        \centering
        \includegraphics[scale=0.2]{supple_figure/BUSI/original_image/val_191_input.png}
        \text{Image}
    \end{minipage}
    \hspace{0.2em}
    \begin{minipage}[t]{0.08\linewidth}
        \centering
        \includegraphics[scale=0.2]{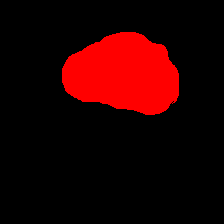}
        \text{GT}
    \end{minipage}
    \hspace{0.2em}
    \begin{minipage}[t]{0.08\linewidth}
        \centering
        \includegraphics[scale=0.2]{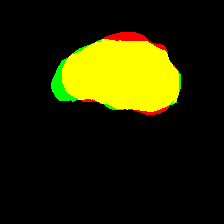}
        \text{CPS}
    \end{minipage}
    \hspace{0.2em}
    \begin{minipage}[t]{0.08\linewidth}
        \centering
        \includegraphics[scale=0.2]{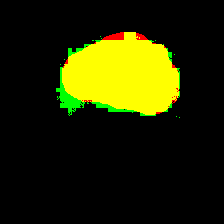}
        \text{ABD}
    \end{minipage}
    \hspace{0.2em}
    \begin{minipage}[t]{0.08\linewidth}
        \centering
        \includegraphics[scale=0.2]{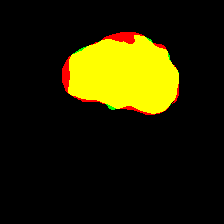}
        \text{AD-MT}
    \end{minipage}
    \hspace{0.2em}
    \begin{minipage}[t]{0.08\linewidth}
        \centering
        \includegraphics[scale=0.2]{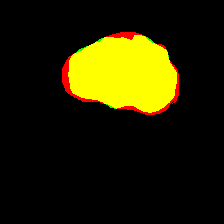}
        \text{DiffRect}
    \end{minipage}
    \hspace{0.2em}
    \begin{minipage}[t]{0.08\linewidth}
        \centering
        \includegraphics[scale=0.2]{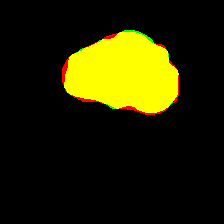}
        \text{DCNet}
    \end{minipage}
    \hspace{0.2em}
    \begin{minipage}[t]{0.08\linewidth}
        \centering
        \includegraphics[scale=0.2]{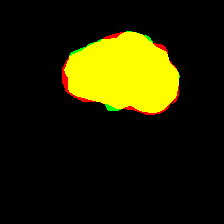}
        \text{SS-Net}
    \end{minipage}
    \hspace{0.2em}
    \begin{minipage}[t]{0.08\linewidth}
        \centering
        \includegraphics[scale=0.2]{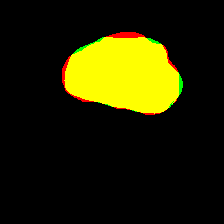}
        \textbf{Hermes(Ours)}
    \end{minipage}
    \caption{Additional visual comparisons of various state-of-the-art methods on the BUSI dataset using 72 labeled images. Red, green, and yellow regions represent ground truth, predictions, and overlapping areas, respectively.}
    \label{sup:vis1}
\end{figure*}

\begin{figure*}[htbp]
    \centering
    \begin{minipage}[t]{0.08\linewidth}   
        \centering
        \includegraphics[scale=0.2]{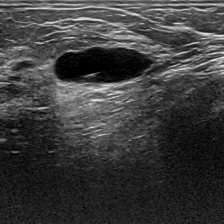}
    \end{minipage}
    \hspace{0.2em}
    \begin{minipage}[t]{0.08\linewidth}
        \centering
        \includegraphics[scale=0.2]{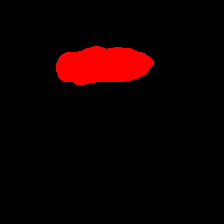}
    \end{minipage}
    \hspace{0.2em}
    \begin{minipage}[t]{0.08\linewidth}
        \centering
        \includegraphics[scale=0.2]{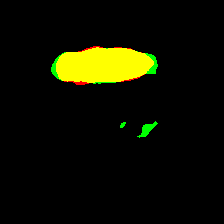}
    \end{minipage}
    \hspace{0.2em}
    \begin{minipage}[t]{0.08\linewidth}
        \centering
        \includegraphics[scale=0.2]{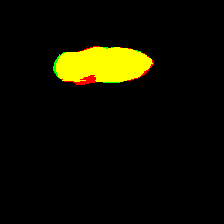}
    \end{minipage}
    \hspace{0.2em}
    \begin{minipage}[t]{0.08\linewidth}
        \centering
        \includegraphics[scale=0.2]{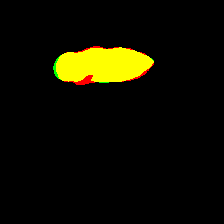}
    \end{minipage}
    \hspace{0.2em}
    \begin{minipage}[t]{0.08\linewidth}
        \centering
        \includegraphics[scale=0.2]{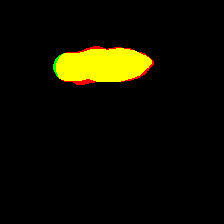}
    \end{minipage}
    \hspace{0.2em}
    \begin{minipage}[t]{0.08\linewidth}
        \centering
        \includegraphics[scale=0.2]{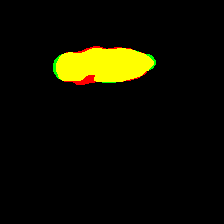}
    \end{minipage}
    \hspace{0.2em}
    \begin{minipage}[t]{0.08\linewidth}
        \centering
        \includegraphics[scale=0.2]{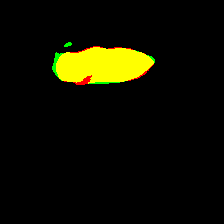}
    \end{minipage}
    \hspace{0.2em}
    \begin{minipage}[t]{0.08\linewidth}
        \centering
        \includegraphics[scale=0.2]{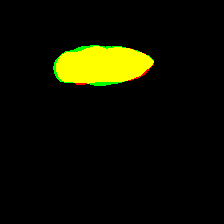}
    \end{minipage}
    \par   
    \vspace{0.1em}
    \begin{minipage}[t]{0.08\linewidth}   
        \centering
        \includegraphics[scale=0.2]{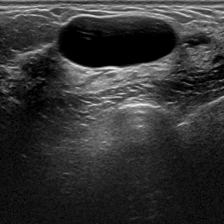}
    \end{minipage}
    \hspace{0.2em}
    \begin{minipage}[t]{0.08\linewidth}
        \centering
        \includegraphics[scale=0.2]{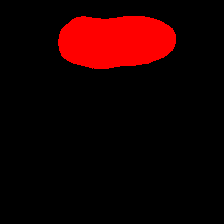}
    \end{minipage}
    \hspace{0.2em}
    \begin{minipage}[t]{0.08\linewidth}
        \centering
        \includegraphics[scale=0.2]{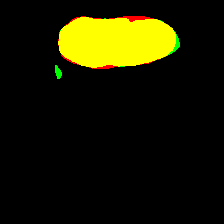}
    \end{minipage}
    \hspace{0.2em}
    \begin{minipage}[t]{0.08\linewidth}
        \centering
        \includegraphics[scale=0.2]{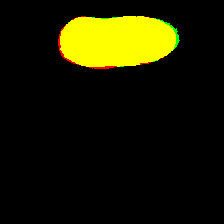}
    \end{minipage}
    \hspace{0.2em}
    \begin{minipage}[t]{0.08\linewidth}
        \centering
        \includegraphics[scale=0.2]{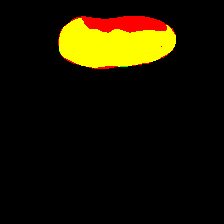}
    \end{minipage}
    \hspace{0.2em}
    \begin{minipage}[t]{0.08\linewidth}
        \centering
        \includegraphics[scale=0.2]{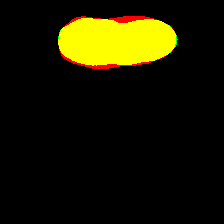}
    \end{minipage}
    \hspace{0.2em}
    \begin{minipage}[t]{0.08\linewidth}
        \centering
        \includegraphics[scale=0.2]{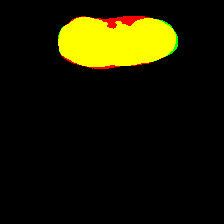}
    \end{minipage}
    \hspace{0.2em}
    \begin{minipage}[t]{0.08\linewidth}
        \centering
        \includegraphics[scale=0.2]{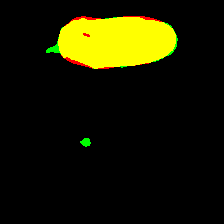}
    \end{minipage}
    \hspace{0.2em}
    \begin{minipage}[t]{0.08\linewidth}
        \centering
        \includegraphics[scale=0.2]{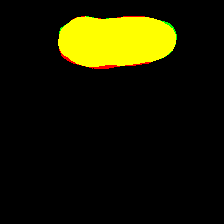}
    \end{minipage}
    \par   
    \vspace{0.1em}
    \begin{minipage}[t]{0.08\linewidth}   
        \centering
        \includegraphics[scale=0.2]{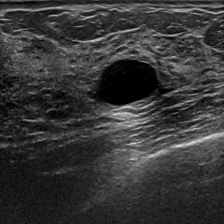}
    \end{minipage}
    \hspace{0.2em}
    \begin{minipage}[t]{0.08\linewidth}
        \centering
        \includegraphics[scale=0.2]{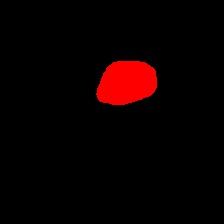}
    \end{minipage}
    \hspace{0.2em}
    \begin{minipage}[t]{0.08\linewidth}
        \centering
        \includegraphics[scale=0.2]{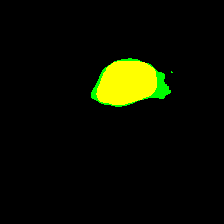}
    \end{minipage}
    \hspace{0.2em}
    \begin{minipage}[t]{0.08\linewidth}
        \centering
        \includegraphics[scale=0.2]{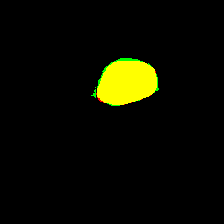}
    \end{minipage}
    \hspace{0.2em}
    \begin{minipage}[t]{0.08\linewidth}
        \centering
        \includegraphics[scale=0.2]{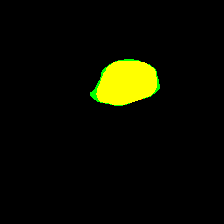}
    \end{minipage}
    \hspace{0.2em}
    \begin{minipage}[t]{0.08\linewidth}
        \centering
        \includegraphics[scale=0.2]{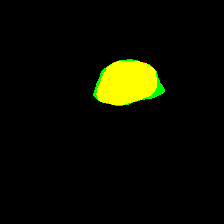}
    \end{minipage}
    \hspace{0.2em}
    \begin{minipage}[t]{0.08\linewidth}
        \centering
        \includegraphics[scale=0.2]{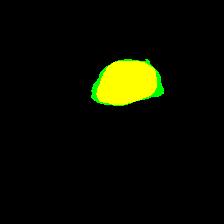}
    \end{minipage}
    \hspace{0.2em}
    \begin{minipage}[t]{0.08\linewidth}
        \centering
        \includegraphics[scale=0.2]{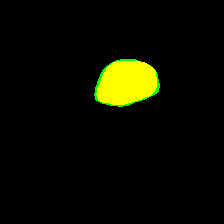}
    \end{minipage}
    \hspace{0.2em}
    \begin{minipage}[t]{0.08\linewidth}
        \centering
        \includegraphics[scale=0.2]{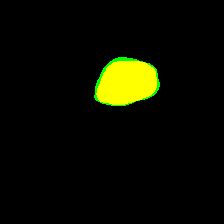}
    \end{minipage}
    \par   
    \vspace{0.1em}
    \begin{minipage}[t]{0.08\linewidth}   
        \centering
        \includegraphics[scale=0.2]{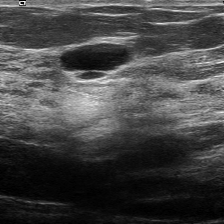}
    \end{minipage}
    \hspace{0.2em}
    \begin{minipage}[t]{0.08\linewidth}
        \centering
        \includegraphics[scale=0.2]{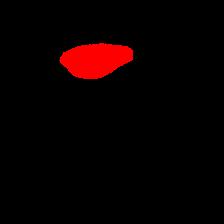}
    \end{minipage}
    \hspace{0.2em}
    \begin{minipage}[t]{0.08\linewidth}
        \centering
        \includegraphics[scale=0.2]{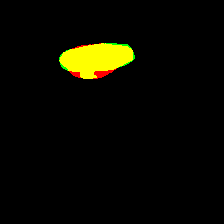}
    \end{minipage}
    \hspace{0.2em}
    \begin{minipage}[t]{0.08\linewidth}
        \centering
        \includegraphics[scale=0.2]{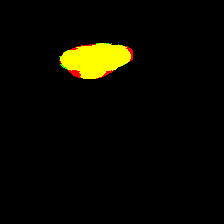}
    \end{minipage}
    \hspace{0.2em}
    \begin{minipage}[t]{0.08\linewidth}
        \centering
        \includegraphics[scale=0.2]{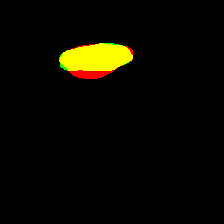}
    \end{minipage}
    \hspace{0.2em}
    \begin{minipage}[t]{0.08\linewidth}
        \centering
        \includegraphics[scale=0.2]{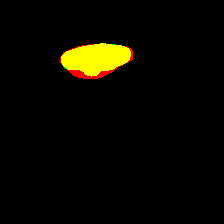}
    \end{minipage}
    \hspace{0.2em}
    \begin{minipage}[t]{0.08\linewidth}
        \centering
        \includegraphics[scale=0.2]{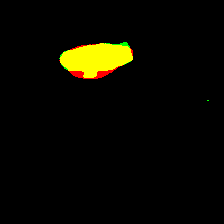}
    \end{minipage}
    \hspace{0.2em}
    \begin{minipage}[t]{0.08\linewidth}
        \centering
        \includegraphics[scale=0.2]{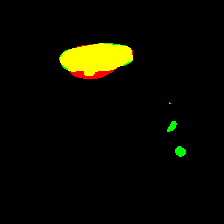}
    \end{minipage}
    \hspace{0.2em}
    \begin{minipage}[t]{0.08\linewidth}
        \centering
        \includegraphics[scale=0.2]{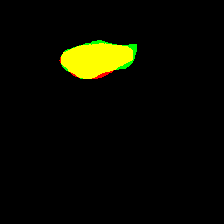}
    \end{minipage}
    \par   
    \vspace{0.1em}
    \begin{minipage}[t]{0.08\linewidth}   
        \centering
        \includegraphics[scale=0.2]{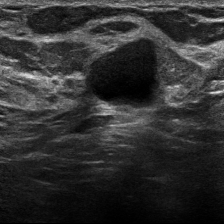}
    \end{minipage}
    \hspace{0.2em}
    \begin{minipage}[t]{0.08\linewidth}
        \centering
        \includegraphics[scale=0.2]{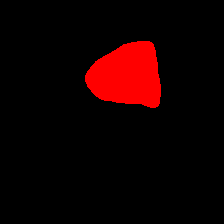}
    \end{minipage}
    \hspace{0.2em}
    \begin{minipage}[t]{0.08\linewidth}
        \centering
        \includegraphics[scale=0.2]{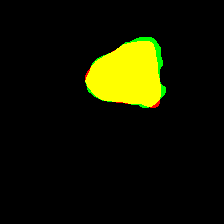}
    \end{minipage}
    \hspace{0.2em}
    \begin{minipage}[t]{0.08\linewidth}
        \centering
        \includegraphics[scale=0.2]{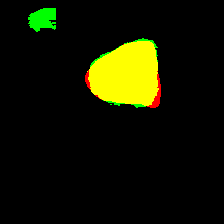}
    \end{minipage}
    \hspace{0.2em}
    \begin{minipage}[t]{0.08\linewidth}
        \centering
        \includegraphics[scale=0.2]{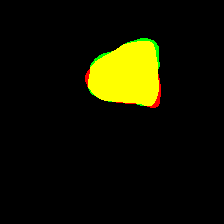}
    \end{minipage}
    \hspace{0.2em}
    \begin{minipage}[t]{0.08\linewidth}
        \centering
        \includegraphics[scale=0.2]{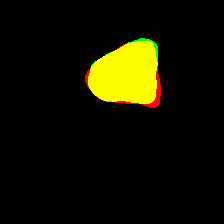}
    \end{minipage}
    \hspace{0.2em}
    \begin{minipage}[t]{0.08\linewidth}
        \centering
        \includegraphics[scale=0.2]{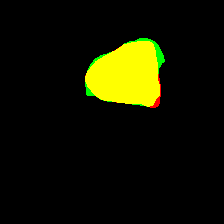}
    \end{minipage}
    \hspace{0.2em}
    \begin{minipage}[t]{0.08\linewidth}
        \centering
        \includegraphics[scale=0.2]{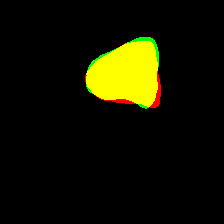}
    \end{minipage}
    \hspace{0.2em}
    \begin{minipage}[t]{0.08\linewidth}
        \centering
        \includegraphics[scale=0.2]{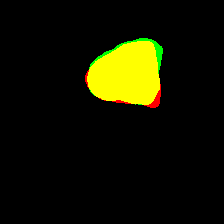}
    \end{minipage}
    \par   
    \vspace{0.1em}
    \begin{minipage}[t]{0.08\linewidth}   
        \centering
        \includegraphics[scale=0.2]{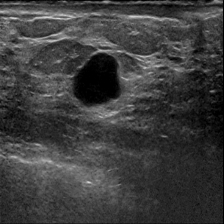}
    \end{minipage}
    \hspace{0.2em}
    \begin{minipage}[t]{0.08\linewidth}
        \centering
        \includegraphics[scale=0.2]{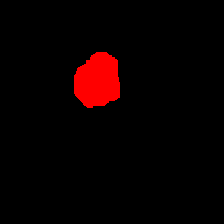}
    \end{minipage}
    \hspace{0.2em}
    \begin{minipage}[t]{0.08\linewidth}
        \centering
        \includegraphics[scale=0.2]{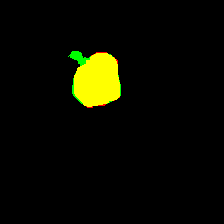}
    \end{minipage}
    \hspace{0.2em}
    \begin{minipage}[t]{0.08\linewidth}
        \centering
        \includegraphics[scale=0.2]{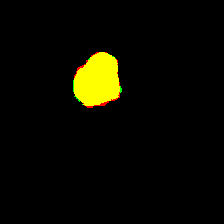}
    \end{minipage}
    \hspace{0.2em}
    \begin{minipage}[t]{0.08\linewidth}
        \centering
        \includegraphics[scale=0.2]{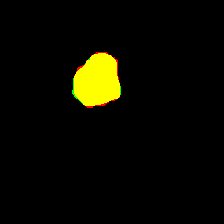}
    \end{minipage}
    \hspace{0.2em}
    \begin{minipage}[t]{0.08\linewidth}
        \centering
        \includegraphics[scale=0.2]{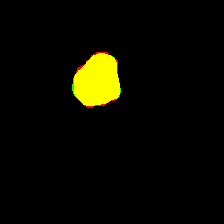}
    \end{minipage}
    \hspace{0.2em}
    \begin{minipage}[t]{0.08\linewidth}
        \centering
        \includegraphics[scale=0.2]{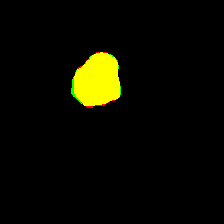}
    \end{minipage}
    \hspace{0.2em}
    \begin{minipage}[t]{0.08\linewidth}
        \centering
        \includegraphics[scale=0.2]{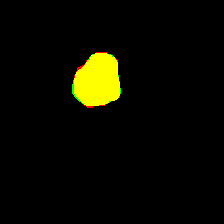}
    \end{minipage}
    \hspace{0.2em}
    \begin{minipage}[t]{0.08\linewidth}
        \centering
        \includegraphics[scale=0.2]{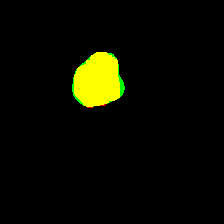}
    \end{minipage}
    \par   
    \vspace{0.1em}
    \begin{minipage}[t]{0.08\linewidth}   
        \centering
        \includegraphics[scale=0.2]{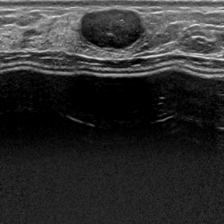}
    \end{minipage}
    \hspace{0.2em}
    \begin{minipage}[t]{0.08\linewidth}
        \centering
        \includegraphics[scale=0.2]{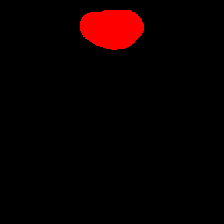}
    \end{minipage}
    \hspace{0.2em}
    \begin{minipage}[t]{0.08\linewidth}
        \centering
        \includegraphics[scale=0.2]{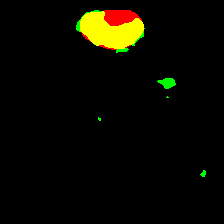}
    \end{minipage}
    \hspace{0.2em}
    \begin{minipage}[t]{0.08\linewidth}
        \centering
        \includegraphics[scale=0.2]{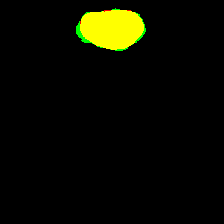}
    \end{minipage}
    \hspace{0.2em}
    \begin{minipage}[t]{0.08\linewidth}
        \centering
        \includegraphics[scale=0.2]{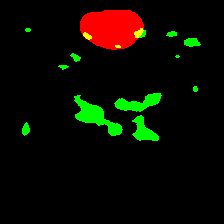}
    \end{minipage}
    \hspace{0.2em}
    \begin{minipage}[t]{0.08\linewidth}
        \centering
        \includegraphics[scale=0.2]{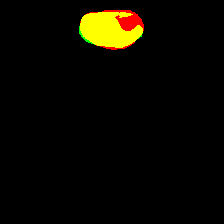}
    \end{minipage}
    \hspace{0.2em}
    \begin{minipage}[t]{0.08\linewidth}
        \centering
        \includegraphics[scale=0.2]{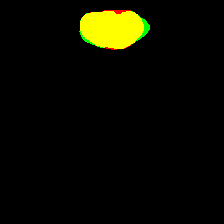}
    \end{minipage}
    \hspace{0.2em}
    \begin{minipage}[t]{0.08\linewidth}
        \centering
        \includegraphics[scale=0.2]{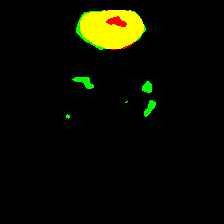}
    \end{minipage}
    \hspace{0.2em}
    \begin{minipage}[t]{0.08\linewidth}
        \centering
        \includegraphics[scale=0.2]{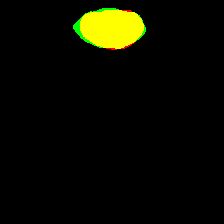}
    \end{minipage}
    \par   
    \vspace{0.1em}
    \begin{minipage}[t]{0.08\linewidth}   
        \centering
        \includegraphics[scale=0.2]{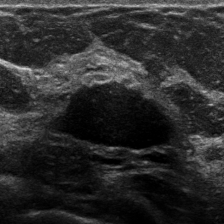}
        \text{Image}
    \end{minipage}
    \hspace{0.2em}
    \begin{minipage}[t]{0.08\linewidth}
        \centering
        \includegraphics[scale=0.2]{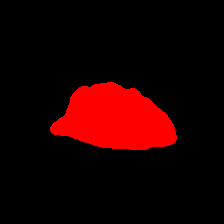}
        \text{GT}
    \end{minipage}
    \hspace{0.2em}
    \begin{minipage}[t]{0.08\linewidth}
        \centering
        \includegraphics[scale=0.2]{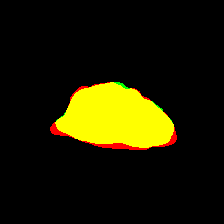}
        \text{CPS}
    \end{minipage}
    \hspace{0.2em}
    \begin{minipage}[t]{0.08\linewidth}
        \centering
        \includegraphics[scale=0.2]{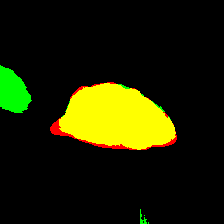}
        \text{ABD}
    \end{minipage}
    \hspace{0.2em}
    \begin{minipage}[t]{0.08\linewidth}
        \centering
        \includegraphics[scale=0.2]{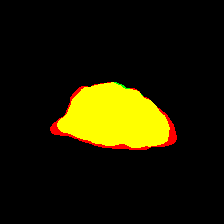}
        \text{AD-MT}
    \end{minipage}
    \hspace{0.2em}
    \begin{minipage}[t]{0.08\linewidth}
        \centering
        \includegraphics[scale=0.2]{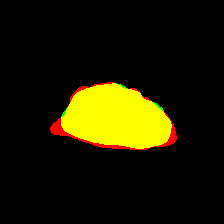}
        \text{DiffRect}
    \end{minipage}
    \hspace{0.2em}
    \begin{minipage}[t]{0.08\linewidth}
        \centering
        \includegraphics[scale=0.2]{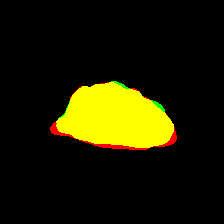}
        \text{DCNet}
    \end{minipage}
    \hspace{0.2em}
    \begin{minipage}[t]{0.08\linewidth}
        \centering
        \includegraphics[scale=0.2]{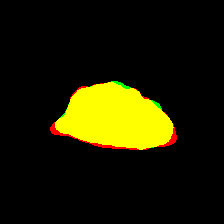}
        \text{SS-Net}
    \end{minipage}
    \hspace{0.2em}
    \begin{minipage}[t]{0.08\linewidth}
        \centering
        \includegraphics[scale=0.2]{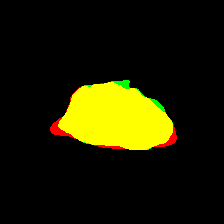}
        \textbf{Hermes(Ours)}
    \end{minipage}
    \caption{Additional visual comparisons of various state-of-the-art methods on the UDIAT dataset using 40 labeled images. Red, green, and yellow regions represent ground truth, predictions, and overlapping areas, respectively.}
    \label{sup:vis2}
\end{figure*}

\begin{figure*}[htbp]
    \centering
    \begin{minipage}[t]{0.08\linewidth}   
        \centering
        \includegraphics[scale=0.2]{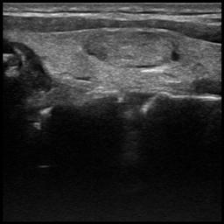}
    \end{minipage}
    \hspace{0.2em}
    \begin{minipage}[t]{0.08\linewidth}
        \centering
        \includegraphics[scale=0.2]{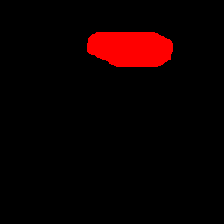}
    \end{minipage}
    \hspace{0.2em}
    \begin{minipage}[t]{0.08\linewidth}
        \centering
        \includegraphics[scale=0.2]{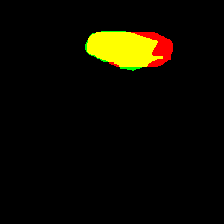}
    \end{minipage}
    \hspace{0.2em}
    \begin{minipage}[t]{0.08\linewidth}
        \centering
        \includegraphics[scale=0.2]{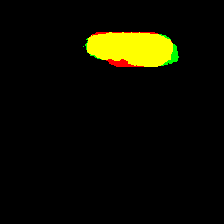}
    \end{minipage}
    \hspace{0.2em}
    \begin{minipage}[t]{0.08\linewidth}
        \centering
        \includegraphics[scale=0.2]{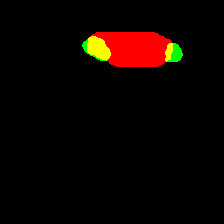}
    \end{minipage}
    \hspace{0.2em}
    \begin{minipage}[t]{0.08\linewidth}
        \centering
        \includegraphics[scale=0.2]{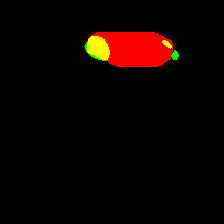}
    \end{minipage}
    \hspace{0.2em}
    \begin{minipage}[t]{0.08\linewidth}
        \centering
        \includegraphics[scale=0.2]{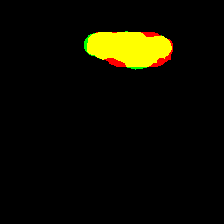}
    \end{minipage}
    \hspace{0.2em}
    \begin{minipage}[t]{0.08\linewidth}
        \centering
        \includegraphics[scale=0.2]{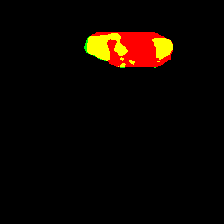}
    \end{minipage}
    \hspace{0.2em}
    \begin{minipage}[t]{0.08\linewidth}
        \centering
        \includegraphics[scale=0.2]{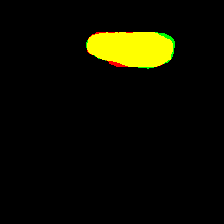}
    \end{minipage}
    \par   
    \vspace{0.1em}
    \begin{minipage}[t]{0.08\linewidth}   
        \centering
        \includegraphics[scale=0.2]{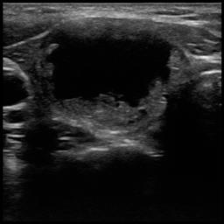}
    \end{minipage}
    \hspace{0.2em}
    \begin{minipage}[t]{0.08\linewidth}
        \centering
        \includegraphics[scale=0.2]{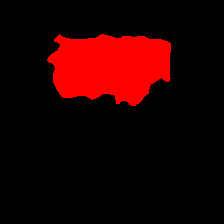}
    \end{minipage}
    \hspace{0.2em}
    \begin{minipage}[t]{0.08\linewidth}
        \centering
        \includegraphics[scale=0.2]{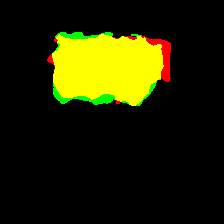}
    \end{minipage}
    \hspace{0.2em}
    \begin{minipage}[t]{0.08\linewidth}
        \centering
        \includegraphics[scale=0.2]{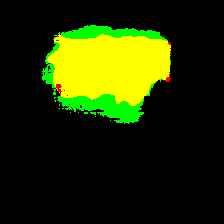}
    \end{minipage}
    \hspace{0.2em}
    \begin{minipage}[t]{0.08\linewidth}
        \centering
        \includegraphics[scale=0.2]{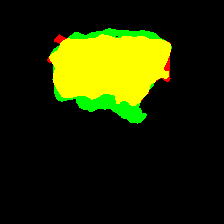}
    \end{minipage}
    \hspace{0.2em}
    \begin{minipage}[t]{0.08\linewidth}
        \centering
        \includegraphics[scale=0.2]{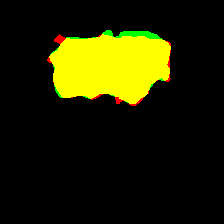}
    \end{minipage}
    \hspace{0.2em}
    \begin{minipage}[t]{0.08\linewidth}
        \centering
        \includegraphics[scale=0.2]{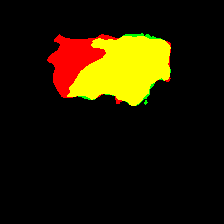}
    \end{minipage}
    \hspace{0.2em}
    \begin{minipage}[t]{0.08\linewidth}
        \centering
        \includegraphics[scale=0.2]{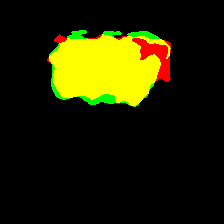}
    \end{minipage}
    \hspace{0.2em}
    \begin{minipage}[t]{0.08\linewidth}
        \centering
        \includegraphics[scale=0.2]{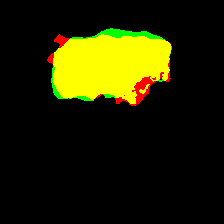}
    \end{minipage}
    \par   
    \vspace{0.1em}
    \begin{minipage}[t]{0.08\linewidth}   
        \centering
        \includegraphics[scale=0.2]{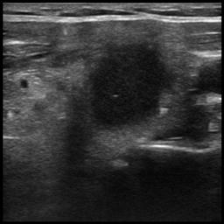}
    \end{minipage}
    \hspace{0.2em}
    \begin{minipage}[t]{0.08\linewidth}
        \centering
        \includegraphics[scale=0.2]{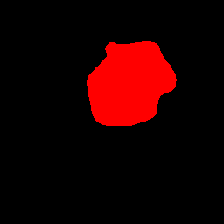}
    \end{minipage}
    \hspace{0.2em}
    \begin{minipage}[t]{0.08\linewidth}
        \centering
        \includegraphics[scale=0.2]{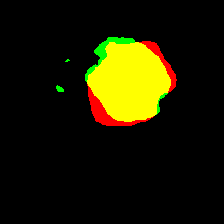}
    \end{minipage}
    \hspace{0.2em}
    \begin{minipage}[t]{0.08\linewidth}
        \centering
        \includegraphics[scale=0.2]{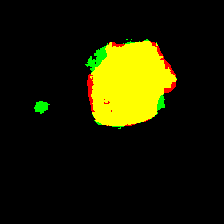}
    \end{minipage}
    \hspace{0.2em}
    \begin{minipage}[t]{0.08\linewidth}
        \centering
        \includegraphics[scale=0.2]{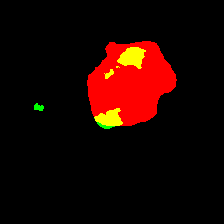}
    \end{minipage}
    \hspace{0.2em}
    \begin{minipage}[t]{0.08\linewidth}
        \centering
        \includegraphics[scale=0.2]{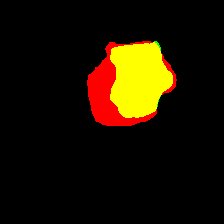}
    \end{minipage}
    \hspace{0.2em}
    \begin{minipage}[t]{0.08\linewidth}
        \centering
        \includegraphics[scale=0.2]{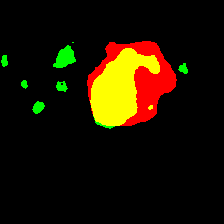}
    \end{minipage}
    \hspace{0.2em}
    \begin{minipage}[t]{0.08\linewidth}
        \centering
        \includegraphics[scale=0.2]{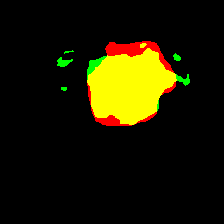}
    \end{minipage}
    \hspace{0.2em}
    \begin{minipage}[t]{0.08\linewidth}
        \centering
        \includegraphics[scale=0.2]{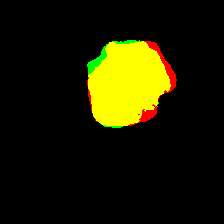}
    \end{minipage}
    \par   
    \vspace{0.1em}
    \begin{minipage}[t]{0.08\linewidth}   
        \centering
        \includegraphics[scale=0.2]{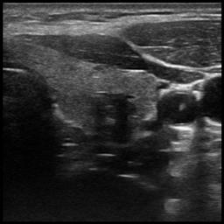}
    \end{minipage}
    \hspace{0.2em}
    \begin{minipage}[t]{0.08\linewidth}
        \centering
        \includegraphics[scale=0.2]{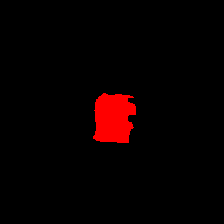}
    \end{minipage}
    \hspace{0.2em}
    \begin{minipage}[t]{0.08\linewidth}
        \centering
        \includegraphics[scale=0.2]{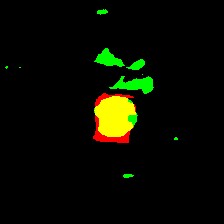}
    \end{minipage}
    \hspace{0.2em}
    \begin{minipage}[t]{0.08\linewidth}
        \centering
        \includegraphics[scale=0.2]{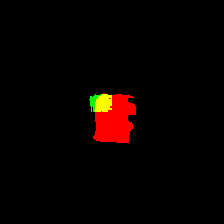}
    \end{minipage}
    \hspace{0.2em}
    \begin{minipage}[t]{0.08\linewidth}
        \centering
        \includegraphics[scale=0.2]{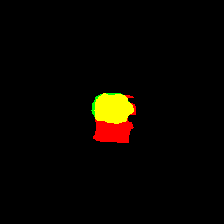}
    \end{minipage}
    \hspace{0.2em}
    \begin{minipage}[t]{0.08\linewidth}
        \centering
        \includegraphics[scale=0.2]{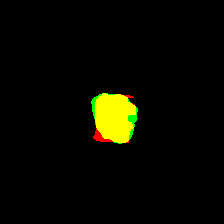}
    \end{minipage}
    \hspace{0.2em}
    \begin{minipage}[t]{0.08\linewidth}
        \centering
        \includegraphics[scale=0.2]{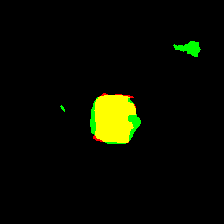}
    \end{minipage}
    \hspace{0.2em}
    \begin{minipage}[t]{0.08\linewidth}
        \centering
        \includegraphics[scale=0.2]{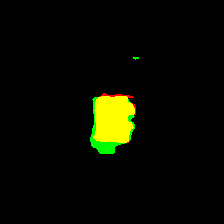}
    \end{minipage}
    \hspace{0.2em}
    \begin{minipage}[t]{0.08\linewidth}
        \centering
        \includegraphics[scale=0.2]{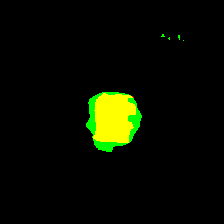}
    \end{minipage}
    \par   
    \vspace{0.1em}
    \begin{minipage}[t]{0.08\linewidth}   
        \centering
        \includegraphics[scale=0.2]{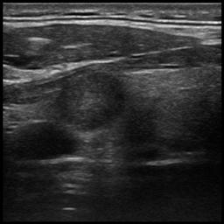}
    \end{minipage}
    \hspace{0.2em}
    \begin{minipage}[t]{0.08\linewidth}
        \centering
        \includegraphics[scale=0.2]{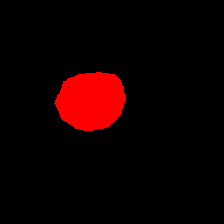}
    \end{minipage}
    \hspace{0.2em}
    \begin{minipage}[t]{0.08\linewidth}
        \centering
        \includegraphics[scale=0.2]{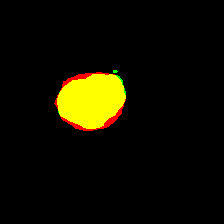}
    \end{minipage}
    \hspace{0.2em}
    \begin{minipage}[t]{0.08\linewidth}
        \centering
        \includegraphics[scale=0.2]{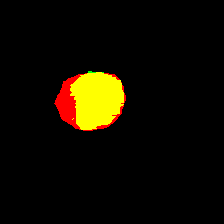}
    \end{minipage}
    \hspace{0.2em}
    \begin{minipage}[t]{0.08\linewidth}
        \centering
        \includegraphics[scale=0.2]{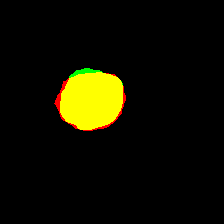}
    \end{minipage}
    \hspace{0.2em}
    \begin{minipage}[t]{0.08\linewidth}
        \centering
        \includegraphics[scale=0.2]{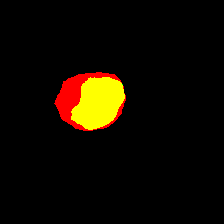}
    \end{minipage}
    \hspace{0.2em}
    \begin{minipage}[t]{0.08\linewidth}
        \centering
        \includegraphics[scale=0.2]{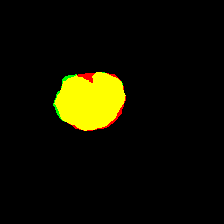}
    \end{minipage}
    \hspace{0.2em}
    \begin{minipage}[t]{0.08\linewidth}
        \centering
        \includegraphics[scale=0.2]{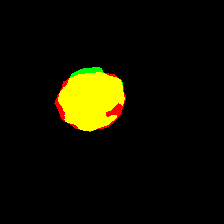}
    \end{minipage}
    \hspace{0.2em}
    \begin{minipage}[t]{0.08\linewidth}
        \centering
        \includegraphics[scale=0.2]{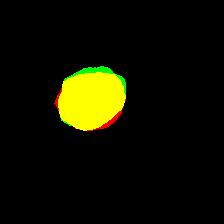}
    \end{minipage}
    \par   
    \vspace{0.1em}
    \begin{minipage}[t]{0.08\linewidth}   
        \centering
        \includegraphics[scale=0.2]{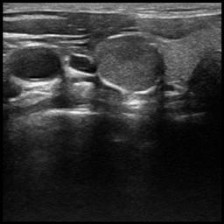}
    \end{minipage}
    \hspace{0.2em}
    \begin{minipage}[t]{0.08\linewidth}
        \centering
        \includegraphics[scale=0.2]{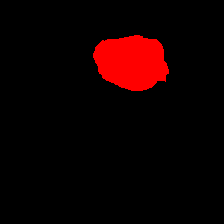}
    \end{minipage}
    \hspace{0.2em}
    \begin{minipage}[t]{0.08\linewidth}
        \centering
        \includegraphics[scale=0.2]{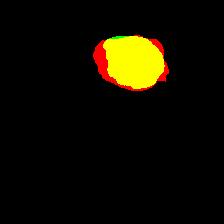}
    \end{minipage}
    \hspace{0.2em}
    \begin{minipage}[t]{0.08\linewidth}
        \centering
        \includegraphics[scale=0.2]{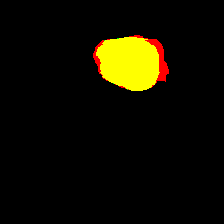}
    \end{minipage}
    \hspace{0.2em}
    \begin{minipage}[t]{0.08\linewidth}
        \centering
        \includegraphics[scale=0.2]{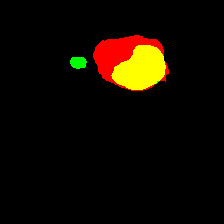}
    \end{minipage}
    \hspace{0.2em}
    \begin{minipage}[t]{0.08\linewidth}
        \centering
        \includegraphics[scale=0.2]{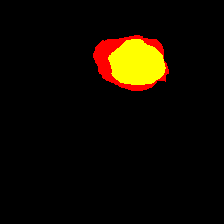}
    \end{minipage}
    \hspace{0.2em}
    \begin{minipage}[t]{0.08\linewidth}
        \centering
        \includegraphics[scale=0.2]{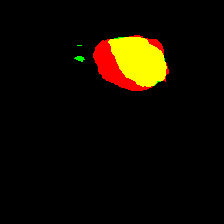}
    \end{minipage}
    \hspace{0.2em}
    \begin{minipage}[t]{0.08\linewidth}
        \centering
        \includegraphics[scale=0.2]{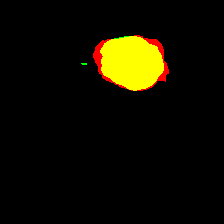}
    \end{minipage}
    \hspace{0.2em}
    \begin{minipage}[t]{0.08\linewidth}
        \centering
        \includegraphics[scale=0.2]{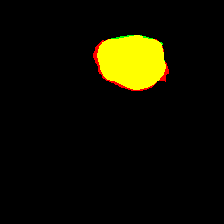}
    \end{minipage}
    \par   
    \vspace{0.1em}
    \begin{minipage}[t]{0.08\linewidth}   
        \centering
        \includegraphics[scale=0.2]{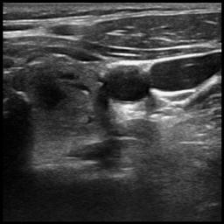}
    \end{minipage}
    \hspace{0.2em}
    \begin{minipage}[t]{0.08\linewidth}
        \centering
        \includegraphics[scale=0.2]{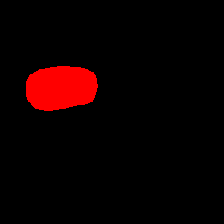}
    \end{minipage}
    \hspace{0.2em}
    \begin{minipage}[t]{0.08\linewidth}
        \centering
        \includegraphics[scale=0.2]{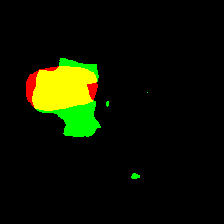}
    \end{minipage}
    \hspace{0.2em}
    \begin{minipage}[t]{0.08\linewidth}
        \centering
        \includegraphics[scale=0.2]{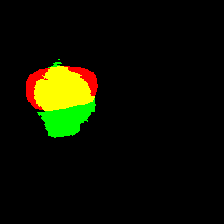}
    \end{minipage}
    \hspace{0.2em}
    \begin{minipage}[t]{0.08\linewidth}
        \centering
        \includegraphics[scale=0.2]{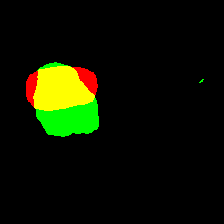}
    \end{minipage}
    \hspace{0.2em}
    \begin{minipage}[t]{0.08\linewidth}
        \centering
        \includegraphics[scale=0.2]{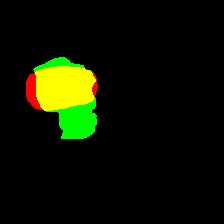}
    \end{minipage}
    \hspace{0.2em}
    \begin{minipage}[t]{0.08\linewidth}
        \centering
        \includegraphics[scale=0.2]{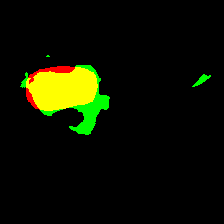}
    \end{minipage}
    \hspace{0.2em}
    \begin{minipage}[t]{0.08\linewidth}
        \centering
        \includegraphics[scale=0.2]{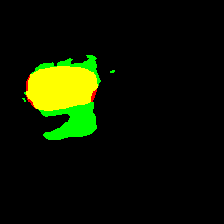}
    \end{minipage}
    \hspace{0.2em}
    \begin{minipage}[t]{0.08\linewidth}
        \centering
        \includegraphics[scale=0.2]{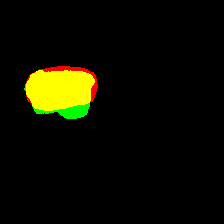}
    \end{minipage}
    \par   
    \vspace{0.1em}
    \begin{minipage}[t]{0.08\linewidth}   
        \centering
        \includegraphics[scale=0.2]{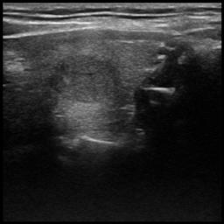}
        \text{Image}
    \end{minipage}
    \hspace{0.2em}
    \begin{minipage}[t]{0.08\linewidth}
        \centering
        \includegraphics[scale=0.2]{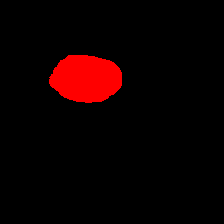}
        \text{GT}
    \end{minipage}
    \hspace{0.2em}
    \begin{minipage}[t]{0.08\linewidth}
        \centering
        \includegraphics[scale=0.2]{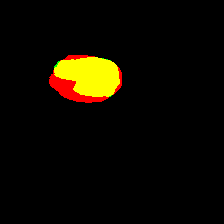}
        \text{CPS}
    \end{minipage}
    \hspace{0.2em}
    \begin{minipage}[t]{0.08\linewidth}
        \centering
        \includegraphics[scale=0.2]{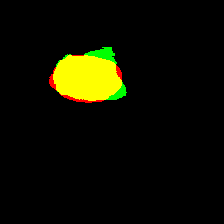}
        \text{ABD}
    \end{minipage}
    \hspace{0.2em}
    \begin{minipage}[t]{0.08\linewidth}
        \centering
        \includegraphics[scale=0.2]{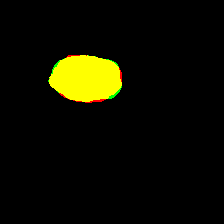}
        \text{AD-MT}
    \end{minipage}
    \hspace{0.2em}
    \begin{minipage}[t]{0.08\linewidth}
        \centering
        \includegraphics[scale=0.2]{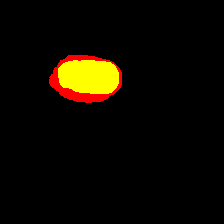}
        \text{DiffRect}
    \end{minipage}
    \hspace{0.2em}
    \begin{minipage}[t]{0.08\linewidth}
        \centering
        \includegraphics[scale=0.2]{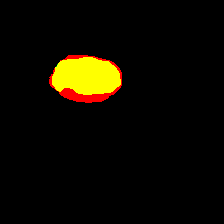}
        \text{DCNet}
    \end{minipage}
    \hspace{0.2em}
    \begin{minipage}[t]{0.08\linewidth}
        \centering
        \includegraphics[scale=0.2]{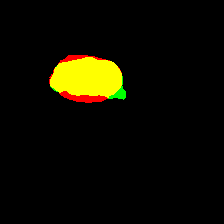}
        \text{SS-Net}
    \end{minipage}
    \hspace{0.2em}
    \begin{minipage}[t]{0.08\linewidth}
        \centering
        \includegraphics[scale=0.2]{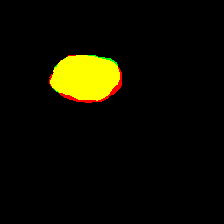}
        \textbf{Hermes(Ours)}
    \end{minipage}
    \caption{Additional visual comparisons of various state-of-the-art methods on the SZ-TUS dataset using 60 labeled images. Red, green, and yellow regions represent ground truth, predictions, and overlapping areas, respectively.}
    \label{sup:vis3}
\end{figure*}

We perform a comprehensive visual comparison against six state-of-the-art methods: CPS [5], ABD [7], AD-MT [39], DiffRect [15], DCNet [3], and SS-Net [29], using U-Net [20] as the common backbone. This evaluation is conducted across three challenging ultrasound datasets—BUSI, UDIAT, and SZ-TUS—to assess the robustness and generalization capability of each method. As shown in Fig. \ref{sup:vis1}, \ref{sup:vis2}, and \ref{sup:vis3}, \textbf{Hermes} consistently demonstrates superior visual results in comparison to the other approaches. The segmentation boundaries are more precise, with better alignment to ground truth annotations, and the predicted lesion areas exhibit fewer false positives and negatives. These improvements are especially obvious in complex and noisy regions, where competing methods often fail to capture fine details or produce overly smooth predictions. This comparison highlights the effectiveness of our dual-branch architecture and the integration of inter-task attention and saliency mechanisms, which together enable more accurate segmentation and classification in ultrasound images.

\end{document}